\documentclass[10pt]{article} 
\usepackage[accepted]{tmlr}

\usepackage{hyperref}
\usepackage{url}

\usepackage{booktabs}       
\usepackage{amsfonts}       
\usepackage{nicefrac}       
\usepackage{microtype}      
\usepackage{xcolor}         
\usepackage{amsmath, multirow, graphicx, caption, subcaption, wrapfig}
\usepackage{floatrow}
\newfloatcommand{capbtabbox}{table}[][\FBwidth]

\newcommand\numberthis{\addtocounter{equation}{1}\tag{\theequation}} 

\usepackage{color, colortbl}
\definecolor{BrownRed}{rgb}{0.8078, 0.2745, 0.2745}
\definecolor{LemonChiffon}{rgb}{1, 0.98, 0.804}
\usepackage{enumerate}
\usepackage[shortlabels]{enumitem}

\usepackage{bm}

\def\eqref#1{equation~(\ref{#1})}

\def\1{\bm{1}}

\def\vb{{\bm{b}}}

\def\ve{{\bm{e}}}

\def\vh{{\bm{h}}}

\def\vu{{\bm{u}}}

\def\vx{{\bm{x}}}

\def\mE{{\bm{E}}}

\def\mH{{\bm{H}}}

\def\mU{{\bm{U}}}

\def\mW{{\bm{W}}}
\def\mX{{\bm{X}}}

\DeclareMathAlphabet{\mathsfit}{\encodingdefault}{\sfdefault}{m}{sl}
\SetMathAlphabet{\mathsfit}{bold}{\encodingdefault}{\sfdefault}{bx}{n}

\def\gD{{\mathcal{D}}}

\def\gS{{\mathcal{S}}}

\def\sR{{\mathbb{R}}}

\usepackage[acronym, nohypertypes={acronym}]{glossaries}

\newacronym{stgnn}{STGNN}{\textit{spatiotemporal graph neural network}}
\newacronym{uq}{UQ}{uncertainty quantification}
\newacronym{gnn}{GNN}{graph neural network}
\newacronym{gdl}{GDL}{graph deep learning}
\newacronym{mp}{MP}{message-passing}
\newacronym{mlp}{MLP}{multilayer perceptron}
\newacronym{tts}{TTS}{time-then-space}
\newacronym{stt}{STT}{space-then-time}
\newacronym{tas}{T\&S}{time-and-space}
\newacronym{rnn}{RNN}{\textit{recurrent neural network}}
\newacronym{narx}{NARX}{nonlinear autoregressive exogenous}
\newacronym{tcn}{TCN}{temporal convolutional network}
\newacronym{gru}{GRU}{gated recurrent unit}
\newacronym{sn}{SN}{\textit{sensor network}}
\newacronym{iot}{IoT}{Internet of Things}
\newacronym{stmp}{STMP}{spatiotemporal message-passing}
\newacronym{stcn}{STCN}{spatiotemporal convolutional network}
\newacronym{gcrnn}{GCRNN}{graph convolutional recurrent neural network}
\newacronym{mae}{MAE}{mean absolute error}
\newacronym{mape}{MAPE}{mean absolute percentage error}
\newacronym{rmse}{RMSE}{root mean squared error}
\newacronym{mre}{MRE}{mean relative error}
\newacronym{mmre}{MMRE}{multivariate mean relative error}
\newacronym{tsp}{TSP}{time series processing}
\newacronym{statt}{STAtt}{\textit{spatiotemporal attention}}
\newacronym{nlp}{NLP}{natural language processing}

\glsdisablehyper

\newglossaryentry{dcrnn}{name=DCRNN,description=}
\newglossaryentry{agcrn}{name=AGCRN,description=}
\newglossaryentry{gwnet}{name=GraphWaveNet,description=}

\newglossaryentry{la}{name=METR-LA,description=}
\newglossaryentry{bay}{name=PEMS-BAY,description=}
\newglossaryentry{cer}{name=CER-E,description=}
\newglossaryentry{air}{name=AQI,description=}
\newglossaryentry{engrad}{name=EngRAD,description=}
\newglossaryentry{climate}{name=CLM-D,description=}
\newglossaryentry{pems3}{name=PEMS03,description=}
\newglossaryentry{pems4}{name=PEMS04,description=}
\newglossaryentry{pems7}{name=PEMS07,description=}
\newglossaryentry{pems8}{name=PEMS08,description=}

\title{On the Regularization of Learnable Embeddings for Time Series Forecasting}

\author{
\name \hspace{-.05in}Luca Butera \email luca.butera@usi.ch \\
\addr Università della Svizzera Italiana, IDSIA \\
\AND
\name Giovanni De Felice \email g.de-felice@liverpool.ac.uk \\
\addr University of Liverpool \\
\AND
\name Andrea Cini \email andrea.cini@usi.ch\\
\addr Università della Svizzera Italiana, IDSIA \\
\AND
\name Cesare Alippi \email cesare.alippi@usi.ch \\
\addr Università della Svizzera Italiana, IDSIA \\
Politecnico di Milano \\
}

\def\month{02}  
\def\year{2025} 
\def\openreview{\url{https://openreview.net/forum?id=F5ALCh3GWG}} 

\begin{document}

\maketitle

\begin{abstract}
    In forecasting multiple time series, accounting for the individual features of each sequence can be challenging. To address this, modern deep learning methods for time series analysis combine a shared~(global) model with local layers, specific to each time series, often implemented as learnable embeddings. Ideally, these \textit{local embeddings} should encode meaningful representations of the unique dynamics of each sequence. However, when these are learned end-to-end as parameters of a forecasting model, they may end up acting as mere sequence identifiers. Shared processing blocks may then become reliant on such identifiers, limiting their transferability to new contexts. In this paper, we address this issue by investigating methods to regularize the learning of local learnable embeddings for time series processing. Specifically, we perform the first extensive empirical study on the subject and show how such regularizations consistently improve performance in widely adopted architectures. Furthermore, we show that methods attempting to prevent the co-adaptation of local and global parameters by means of embeddings perturbation are particularly effective in this context. In this regard, we include in the comparison several perturbation-based regularization methods, going as far as periodically resetting the embeddings during training. The obtained results provide an important contribution to understanding the interplay between learnable local parameters and shared processing layers: a key challenge in modern time series processing models and a step toward developing effective foundation models for time series.
\end{abstract}

\section{Introduction}\label{sec:introduction}
Collections of related time series characterize many applications of learning systems in the real world, such as traffic monitoring~\citep{li2018diffusion,yu2018spatio}, energy analytics~\citep{dimoulkas2019neural, gasparin2022deep}, climate modeling~\citep{ma2023histgnn, chen2023foundation}, and biomedical data processing~\citep{jarrett2021clairvoyance, zhang2022graph}. The success of deep learning in the associated tasks, e.g., forecasting~\citep{shih2019temporal,benidis2022deep}, imputation~\citep{cao2018brits,cini2022filling} and virtual sensing~\citep{wu2021inductive}, relies on effectively modeling shared patterns across time series while also accounting for their individual characteristics~\citep{benidis2022deep}. In this context, models must rely on some attributes or positional encodings to tailor the processing to the target time series, with the risk of requiring long observation windows and high model capacity when those are not available~\citep{salinas2020deepar, montero2021principles}. Indeed, prior and positional information is often unavailable or insufficient to provide effective specialization. Notably, the study of methods to tune a shared time series model on a specific task is a major concern for the development of foundation models, i.e., backbone models trained on large collections of time series and then applied to specific target applications~\citep{garza2023timegpt, liang2024foundation}. This problem is also particularly relevant in spatiotemporal forecasting when dealing with multiple time series coming from a sensor network~\citep{bai2020adaptive, cini2023taming}. 

Among different methods, research has addressed the problem by looking into \textit{hybrid} global-local architectures, i.e., architectures that combine global models with local learnable components~(e.g., layers) specific to a target time series~\citep{wang2019deep, smyl2020hybrid}. As a prominent example, \citet{smyl2020hybrid} won the M4 competition~\citep{makridakis2020m4} by combining a shared \gls{rnn} with local exponential smoothing models fitted on each time series. In the context of spatiotemporal data, research has shifted towards the adoption of learnable embeddings~\citep{bai2020adaptive,shao2022spatial}, i.e., vectors of learnable parameters, to reduce the cost of learning~(more complex) local processing blocks~\citep{cini2023taming}.  Each embedding, associated with a specific time series, is fed into the shared modeling architecture and trained end-to-end alongside it. These representations can go beyond simply encoding coordinates, as they can account for the dynamics of each time series with respect to the other sequences in the collection~\citep{cini2023taming}. Parameters of this kind are analogous to the word embeddings used in \gls{nlp}~\citep{mikolov2013distributed, peng2015comparative}.  We will refer to them as \textit{local embeddings}. 

While local embeddings offer significant advantages, their adoption introduces potential drawbacks. In the first place, differently from \gls{nlp} applications that operate on a fixed-size dictionary, the target time series might change over time, and new sequences might be added to the collection. Secondly, since embeddings are learned jointly with the entire architecture, co-adaptation~\citep{srivastava2014dropout} with the shared layers might result in embeddings being used as simple sequence identifiers~\citep{geirhos2020shortcut}. This interdependence can hinder the flexibility of the shared processing blocks and limit their transferability. Existing works~\citep{yin2022nodetrans, cini2023taming, prabowo2024traffic} show evidence that constraining the structure of the embedding space can lead to transferability improvements. However, no prior work has systematically addressed and evaluated regularization methods for local embeddings in forecasting architectures. 

To fill this void, we investigate the impact of regularizing the learning of local embeddings for related time series within a selection of commonly used deep learning forecasting architectures. 
In particular, we perform an extensive empirical study comparing a variety of regularization strategies, ranging from standard approaches, such as weight penalties and dropout, to more advanced methods, e.g., clustering and variational regularization. 
Our analysis considers a range of scenarios, including transductive settings, transfer learning, and sensitivity analyses. Empirical results show that methods that attempt to prevent the co-adaptation of the local and global blocks by perturbing the embeddings at training time are consistently among the best-performing approaches. To further validate this observation, we include in the analysis a \textit{forget-and-relearn} strategy~\citep{zhou2021fortuitous}, named \textit{forgetting regularization}, that periodically resets embeddings during training. The results obtained by considering this approach further show that perturbation of local parameters offers a good design principle for regularization strategies in this context. 
Our main findings can be summarized as follows.
\begin{enumerate}[start=1,label={(\bfseries F\arabic*)}, leftmargin=1.5em, itemindent=1em, itemsep=0cm]
    \item \label{F1} In both transductive and transfer learning settings, regularizing the local embeddings can provide consistent performance improvements across several forecasting architectures.
    \item \label{F2} Regularization strategies based on local parameters' perturbations aimed at preventing co-adaptation consistently lead to larger performance gains compared to other approaches.
    \item \label{F3} Finding \ref{F2} can be used as a design principle for designing new regularization strategies to mitigate overfitting and improve transferability, as highlighted by the competitive performance of the considered \textit{forgetting} regularization.
\end{enumerate}
The regularization of local learnable embeddings emerges as an often neglected but central aspect, which, with negligible computational overhead, can lead to performance improvements across all the considered benchmarks. This makes our study an important missing piece for guiding practitioners and researchers in building and designing modern neural architectures for time series processing.

\section{Preliminaries}

This section introduces the notation and formalizes the problem of time series forecasting, with a focus on the class of deep learning architectures we consider in this paper.

\subsection{Problem settings}

Consider a collection $\gD$ of $N$ time series, where each $i$-th time series consists of a sequence of $T$ observations $\{\vx^i_t \in \sR^{d_x}\}_{t=1}^{T}$. Specifically, we indicate as \textit{related time series} a set of homogenous time series coming from the same domain but generated by different sources~(e.g., different sensors). Examples include sales figures for different products, or energy consumption of various users.  
Time series might be acquired both asynchronously and/or synchronously; in the latter case, we denote the stacked $N$ observations at time step $t$ by the matrix $\mX_t \in \sR^{N \times d_x}$. We use the shorthands $\mX_{t:t+T}$ to indicate the sequence of observations within the time interval $[t, t+T)$ and $\mX_{\leq t}$ to indicate those up to time $t$ (included). Eventual (exogenous) covariates associated with each time series are denoted by $\vu^i_t \in \sR^{d_u}$~($\mU_t \in \sR^{N\times d_u}$). 
We focus on multistep-ahead time series forecasting, i.e., the problem of predicting the next $H$ future values for each $i$-th time series $\vx^i_{t:t+H}$ given exogenous variables and a window of $W$ past observations. We focus on point forecasts, while probabilistic predictors might also be considered.

\subsection{Hybrid global-local architectures for time series} \label{sec:st_models}

We consider a broad class of models similar to those in \citep{benidis2022deep} and \citep{cini2023graph}. In particular, we consider predictors such as
\begin{equation}\label{eq:forecasting-siso}
    \hat\vx^i_{t:t+H} = F(\vx^i_{t-W:t}, \vu^i_{t-W:t+H}; \, \boldsymbol{\theta}), \quad i=1,...,N
\end{equation}
where $\boldsymbol{\theta}$ are the learnable parameters of the model, $\hat\vx^i_{t:t+H}$ indicates the predicted values of $\vx^i_{t:t+H}$, and $F(\cdot;\boldsymbol{\theta})$ is a family of parametric forecasting models. Models in Eq.~\ref{eq:forecasting-siso} do not take into account (spatial) dependencies that might exist among time series. In scenarios such as spatiotemporal forecasting, where such dependencies might be relevant to achieve accurate predictions, models that can operate on sets of (synchronous) correlated time series should be preferred. In particular, we consider a family of models 
\begin{equation}\label{eq:forecasting-mimo}
    \widehat \mX^{\gS}_{t:t+H} = F(\mX^{\gS}_{t-W:t}, \mathbf{U}^{\gS}_{t-W:t+H}; \, \boldsymbol{\theta}), \quad\quad \forall \gS \subseteq \gD
\end{equation}
where $F(\cdot;\boldsymbol{\theta})$ operates on subsets $\gS$ of the collection of time series $\gD$, and $\mX_t^{\gS}\in \sR^{|\gS|\times d_x}$ is the corresponding stack of observations at time step $t$. Models of this kind can be implemented by architectures operating on sets~\citep{zaheer2017deep}, e.g., attention-based architectures~\citep{vaswani2017attention, grigsby2021longrange} or \glspl{stgnn}~\citep{jin2023survey, cini2023graph} based on message passing~\citep{gilmer2017neural}. 
These models can eventually account for priors on existing dependencies among time series, which could be encoded, e.g., by a graph adjacency matrix $\mathbf{A} \in \sR^{N \times N}$~\citep{cini2023graph}. 
Note that models in both families~(Eqs.~\ref{eq:forecasting-siso} and \ref{eq:forecasting-mimo}) are \textit{global}, i.e., they share parameters among all the time series being processed, with clear advantages in terms of sample efficiency~\citep{montero2021principles}. Most of the recent successes in applying deep learning to time series forecasting are based on the idea of implementing such global models with a neural network~\citep{benidis2022deep}. In the following, we will consider models belonging to the family in Eq.~\ref{eq:forecasting-mimo}, as Eq.~\ref{eq:forecasting-siso} can be seen as a special case where $|\gS| = 1$. In particular, we consider the whole collection at once~(i.e., $|\gS| = N$) and drop the superscript $\gS$ to simplify the notation.

\begin{figure}[t]
     \centering
      \includegraphics[width=\textwidth]{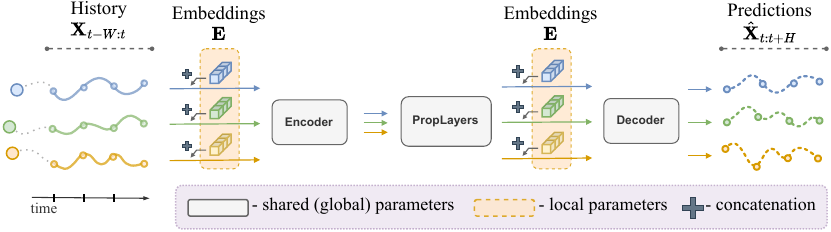}
    \vspace{-.5cm}
    \caption{Overview of the hybrid global-local time series forecasting framework.}
    \label{fig:overview}
\end{figure}

\vspace{0.36em}
\paragraph{Template architecture}

We use a template architecture analogous to that of \citet{cini2023taming}. It consists of the following three processing steps: an \textit{encoder}, one or more \textit{propagation layers}, and a \textit{decoder}. Predictions at each time step and time series are obtained as
\begin{align*}
& \vh_t^{i,0}  = \textsc{Encoder}\left(\vx_{t-1}^i, \, \vu_{t-1}^i\right) \numberthis \label{eq:encoder} \\ 
& \mH_t^{l+1}  = \textsc{PropLayer}^l\left(\mH_{\leq t}^l\right),  \quad l=0,1,...,L-1 \numberthis \label{eq:stp} \\
& \mathbf{\hat{x}}_{t:t+H}^i  = \textsc{Decoder}\left(\vh_{t}^{i,L}, \vu_{t:t+H}\right) \numberthis \label{eq:decoder}
\end{align*}
where the additional index $l$ refers to the layer depth, $\vh^{i,l}_t \in \sR^{d_h}$ is the representation associated with the $i$-th time series at the $l$-th layer, and $\mH^l_t \in \sR^{|\gS|\times d_h}$ the stack of such representations. The $\textsc{Encoder}\left(\cdot\right)$ and $\textsc{Decoder}\left(\cdot\right)$ blocks can be implemented in different ways, e.g. a linear layer or a \gls{mlp}, and operate on single time steps and time series. Conversely, the $\textsc{PropLayer}\left(\cdot\right)$ blocks are the only components of the architecture propagating representations across the temporal dimension and across time series in the collection. Each $\textsc{PropLayer}\left(\cdot\right)$ can be implemented by any existing sequence modeling architecture, e.g., \glspl{rnn}~\citep{hochreiter1997long, cho2014properties} or \glspl{tcn}~\citep{lecun1998convolutional, bai2018empirical}, and/or spatiotemporal operators, e.g.,  \gls{statt} models~\citep{grigsby2021longrange,deihim2023sttre}. In particular, \glspl{stgnn}~\citep{seo2018structured, yu2018spatio}, i.e., models that combine sequence modeling architectures with message passing operators, are among the most popular architectures for implementing propagation layers when processing correlated time series~\citep{jin2023survey}. 

\paragraph{Hybrid global-local architectures and local embeddings}
As anticipated in the introduction, despite their advantages, global models might struggle to account for the specific dynamics of each time series. Hybrid global-local architectures~\citep{benidis2022deep} address this issue by including parameters specific to each target time series. If we indicate these parameters as $\Phi = \{\boldsymbol{\phi}^1, ..., \boldsymbol{\phi}^N\}$, the resulting model family would provide forecasts for the time series in the collection as
\begin{equation}\label{eq:hybrid-forecasting-mimo}
    \widehat \mX^\gS_{t:t+H} = F\left(\mX^\gS_{t-W:t}, \mU^\gS_{t-W:t+H}; \, \boldsymbol{\theta}, \Phi^\gS\right).
\end{equation}
Although many implementations of such models exist~\citep{benidis2022deep, cini2023taming}, we consider models where the local components are implemented as embeddings $\mE \in \sR^{N\times d_e}$ of learnable parameters such that $\Phi = \mE$ and $\boldsymbol{\phi}^i = \ve^i$. In particular, each local embedding $\mathbf{e}^i$ is associated with the corresponding $i$-th time series and can be learned end-to-end jointly with the shared network weights. Embeddings are incorporated into the template architecture at both the encoder and decoder level by concatenating them to the input as 
\begin{equation}
\vh_t^{i,0} = \textsc{Encoder}\left(\vx_{t-1}^i \Vert \, \mathbf{e}^i, \vu_{t-1}^i\right), \quad
\mathbf{\hat{x}}_{t:t+H}^i = \textsc{Decoder}\left(\vh_{t}^{i,L} \Vert \, \mathbf{e}^i, \vu_{t:t+H}\right). \label{eq:encoder_decoder_emb} 
\end{equation}
Fig.~\ref{fig:overview} provides an overview of the resulting reference architecture.
The addition of these parameters comes at a cost in terms of flexibility, as processing time series that were not observed at training time requires fitting new parameters~\citep{januschowski2020criteria, cini2023taming}. 

\section{Related works}
Learnable embeddings are key components of state-of-the-art time series processing architectures such as \glspl{stgnn}~\citep{bai2020adaptive, cini2023graph} and attention-based models~\citep{marisca2022learning, liu2023spatio, xiao2023gaformer}. In particular, \citet{cini2023taming} systematically addresses the role of such embeddings in hybrid global-local \glspl{stgnn}. Aside from modeling local dynamics, embeddings have been used extensively to amortize the cost of learning the full adjacency matrix in graph-based models~\citep{wu2019graph, shang2021discrete, satorras2022multivariate, de2024graph}, or as spatial positional encodings~\citep{marisca2022learning, shao2022spatial, liu2023spatio}. They are also routinely used in \gls{nlp} as word embeddings~\citep{mikolov2013distributed} and often as spatial positional encodings~\citep{devlin2019bert, yang2019xlnet}. 

Methods to regularize learning architectures are obviously central to deep and machine learning in general~\citep{ying2019overview, tian2022comprehensive}. Traditional examples include L1~(lasso)~\citep{tibshirani1996regression} and L2~(ridge, weight decay)~\citep{krogh1991simple} regularizations. Established approaches in deep learning encompass \textit{dropout}~\citep{srivastava2014dropout}, \textit{layer normalization}~\citep{ba2016layer} and \textit{weight normalization}~\citep{salimans2016weight}. Notably, several regularizations have been tailored to specific architectures~\citep{zaremba2014recurrent, gal2016theoretically, dieng2018noisin, wang2019state, santos2022avoiding}, and designed to improve transferability~\citep{wang2018dual, takada2020transfer, abuduweili2021adaptive}. Similarly in spirit to our work, \citet{peng2015comparative} presents a comparative study of different regularizations for word embeddings in \gls{nlp}. Regarding regularizations for learnable embeddings for time series, \citet{yin2022nodetrans} uses learned embedding clusters to regularize fine-tuning on target data, facilitating transfer in graph-based forecasting. Similarly, \citet{cini2023taming} uses a cluster-based regularization, as well as a variational-based approach. However, these works only address model transferability, without considering the impact of local embedding regularization in a broader context, e.g., transductive settings. 

\section{Regularization strategies for local embeddings} \label{sec:regularizations}
In this section, we discuss some possible shortcomings of hybrid global-local forecasting architectures and present the regularization strategies employed in our experimental analysis. 

\begin{wrapfigure}[24]{r}{0.35\textwidth}
    \includegraphics[width=\textwidth]{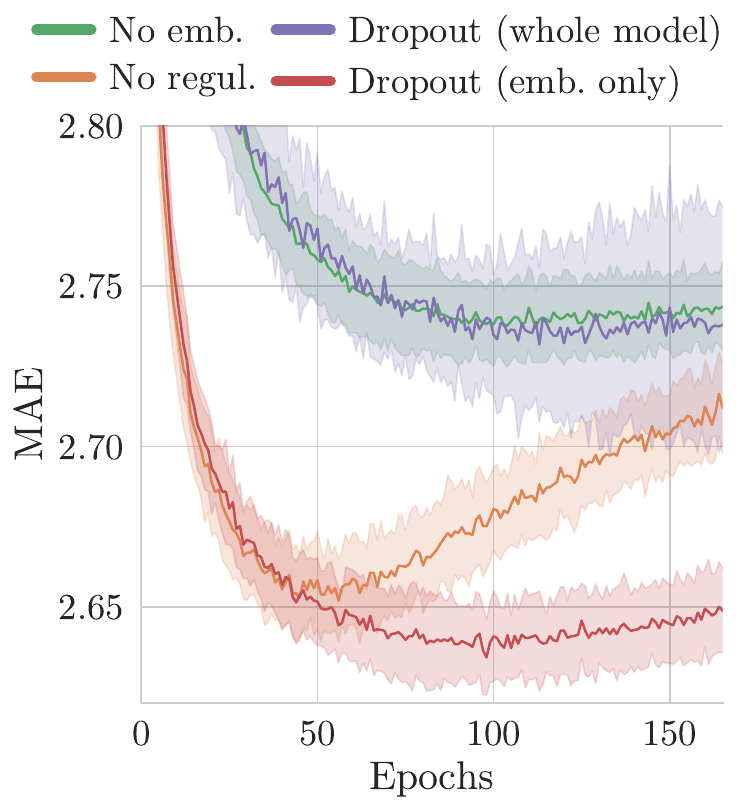}
    \caption{Validation curves when regularizing whole model or just local parameters in a time series forecasting task (STGNN model, METR-LA dataset, 50 runs, $\pm 1std$). Models and datasets are discussed in Sec.~\ref{sec:exp}.}
    \label{fig:dropout_local_global}
\end{wrapfigure}
Global models are inherently less likely to overspecialize to individual sequences since all parameters are shared among time series. The introduction of local embeddings reduces the regularization effect brought by this inductive bias and jointly learning local and global parameters end-to-end on a downstream task may result in a model that is more likely to overfit to individual time series, with a potential negative impact on both performance and transferability~\citep{cini2023taming}. In the following, we use the term \textit{co-adaptation} to refer to these overfitting phenomena coming from training the local embeddings end-to-end with the global model. 

Model regularization is a common approach to deal with overfitting and to control model and sample complexity. For instance, one might apply well-known techniques, such as \textit{weight decay}~\citep{krogh1991simple} or \textit{dropout}~\citep{srivastava2014dropout}, across the entire network. 
However, in many scenarios, the global processing blocks are not particularly overparametrized, as weights are shared, and regularization might not be necessary. In global-local models, one could consider regularizing the local parameters only, i.e., in our settings, constraining how embeddings are learned. 
Fig.~\ref{fig:dropout_local_global} empirically supports this intuition with an illustrative example: given a global time-series forecasting model with fixed hyperparameters~(\textit{green}), adding local embeddings~(\textit{orange}) improves performance, but makes it more likely to overfit~(as model complexity increases). 
Regularizing the entire global-local model~(\textit{purple}), however, hinders performance in this scenario. 
Conversely, applying regularization (dropout in this case) to the embeddings only~(\textit{red}), regularizes the model effectively. 
Note that regularization of the global parameters can obviously still be beneficial in the case of an over-parametrized global model.

\subsection{Regularization methods}\label{subsec:local-regularizations}

In the following, we present the regularization methods that will later be part of our experimental analysis in Sec.~\ref{sec:exp}. They range from adaptations of standard approaches to recent contributions from the literature. We also consider an approach based on parameter resetting.

\paragraph{L1 and L2 regularization} 
L2 regularization~\citep{krogh1991simple}, also known as \textit{weight decay}, consists in adding to the loss a penalty term proportional to the magnitude of the model's weights. When applied to the embeddings, the penalty term is $\mathcal{L}_{l2}\left(\mE\right) = \lambda_{l2} \cdot \|\mE\|_2^2$. Similarly, L1 regularization~\citep{tibshirani1996regression}, also known as \textit{lasso}, consists in applying a penalty term of $\mathcal{L}_{l1}\left(\mE\right) = \lambda_{l1} \cdot \|\mE\|_1$. The positive scalar values, $\lambda_{l2}$ and $\lambda_{l1}$, control the regularization strength.

\paragraph{Dropout}
\textit{Dropout}~\citep{srivastava2014dropout} is another widely recognized regularization. This consists in randomly masking out individual neurons during training, with probability $p$, while scaling the others by $\frac{1}{1 - p}$. This technique heuristically forces the network to learn robust representations, often preventing overfitting. In our context, we apply it to the embedding vectors by randomly masking out some parameters for each embedding. 

\paragraph{Clustering}
The clustering regularization introduced in~\citep{cini2023taming} is based on learning a set of cluster centroids and a cluster assignment matrix, alongside the embeddings. A regularization term, with weight $\lambda_{clst}$, is then added to the loss to minimize the distance between embeddings and the assigned centroid. It was originally designed to structure the embedding space and improve model transferability.

\paragraph{Variational regularization}
\textit{Variational regularization}~\citep{cini2023taming}, consists in modeling embeddings as samples from a Gaussian posterior learned from data $\mathbf{e}^i \sim \mathcal{N}\left(\bm{\mu}^i_{var}, diag\left(\bm{\sigma}^i_{var}\right)\right)$, where $\bm{\mu}^i_{var}$ and $\bm{\sigma}^i_{var}$ are the learnable local parameters. A penalty term based on the KL-divergence between the learned distribution and a standard Gaussian prior, with weight $\lambda_{var}$, is added to the loss. Similarly to the \textit{clustering} regularization, it was introduced to enhance model transferability.

\paragraph{Forgetting}
We also consider a novel \textit{forgetting regularization}, based on the \textit{forget-and-relearn} paradigm~\citep{zhou2021fortuitous}. This refers to strategies where the parameters of some neural network layers are occasionally reset during training. By doing this, it is possible to reduce memorization of training samples~\citep{baldock2021deep}, avoiding shortcuts and learning representations that generalize better~\citep{geirhos2020shortcut,zhou2021fortuitous}. 
We adopt this paradigm in the context of hybrid global-local time series forecasting architectures and propose the following procedure. Every $K$ training epochs, we periodically reset local embeddings $\mE$ to a sample from a shared initialization distribution $\mathcal{P}_e$. 
Concurrently, we similarly reset the $\textsc{Encoder}$'s~ and $\textsc{Decoder}$'s weights~(Eq.~\ref{eq:encoder_decoder_emb}) that directly multiply the embeddings (see Appendix~\ref{app:encoder_decoder_reset}). The value of $K$ can be easily selected empirically (see Appendix~\ref{app:forgetting_sensitivity}). Additionally, resetting is halted after a certain amount of epochs to allow for convergence to a final configuration; this can be easily tuned manually, by considering the model's convergence speed or triggered automatically, by monitoring the validation loss. Note that resetting local parameters to regularize the training of a related time series forecasting model has never been explored in the literature. 

\subsection{Preventing co-adaptation}

As mentioned in Sec.~\ref{sec:regularizations}, learning local and global parameters jointly on a downstream task can result in overfitting. We use the term \textit{co-adaptation} to indicate this phenomenon, by referring to global and local parameters \textit{co-adapting} during training. 
Specifically, local embeddings might simply act as sequence identifiers and result in models that rely entirely on this identification mechanism. Besides the negative effects on sample efficiency, the resulting architecture would likely lose flexibility in terms of the transferability of the learned representations and processing blocks. Furthermore, this behavior is in contrast with the intended one for local embeddings, i.e., encoding meaningful characteristics of the target time series. One way in which regularization methods can aim to prevent such co-adaptation is to actively perturb the local parameters during training so that subsequent layers cannot rely on specific feature values~\citep{srivastava2014dropout}. In our case, the global processing block would be less likely to rely on specific values of the embeddings when processing the target time series. In the following, we will use the term perturbation in a broad sense, referring to strategies that modify the values of local parameters at training time, e.g., resampling, zeroing, or adding noise. 

Among the considered regularization methods, dropout perturbs the embeddings by randomly zeroing out parameters, as similarly done for the inputs/outputs of subsequent layers to avoid the co-adaptation of the associated weights~\citep{srivastava2014dropout}. 
The variational regularization, given a common prior, learns each embedding's sampling distribution end-to-end while resampling their actual values before providing them as inputs to the downstream at each forward pass. 
Finally, forgetting regularization introduces perturbation in the early stages of training by periodically resetting the local parameters. In particular, by considering this strategy for the first time in this context, we aim to test whether actively perturbing local parameters offers an effective principle for the design of new regularizations for hybrid time series forecasting models. 

Regarding the other regularization methods considered in the analysis, L1 and L2 regularizations simply penalize the embeddings' magnitude, while clustering similarly penalizes the distance of the embeddings from the learned centroids, forcing them to occupy specific regions within the embedding space. While these regularizations can provide structure to the embedding space, they do not actively perturb embeddings' values. The empirical results presented in Sec.~\ref{sec:exp} altogether suggest that strategies based on embedding perturbations are more effective at preventing the kind of overfitting we previously discussed, i.e., co-adaptation. 

\section{Experiments}\label{sec:exp}
We evaluate the effectiveness of different regularization strategies for local embeddings under three different scenarios: time series forecasting benchmarks~(Sec.~\ref{subsec:benchmarks}), transfer learning~(Sec.~\ref{subsec:transfer}), and a sensitivity analysis through embedding perturbations~(Sec.~\ref{subsec:perturbation}).  
We consider six real-world datasets of time series collections, spanning four different application domains: \textbf{\gls{la}} and \textbf{\gls{bay}}~\citep{li2018diffusion} as two established benchmarks for traffic forecasting, 
\textbf{\gls{air}}~\citep{zheng2015forecasting} from the air quality domain, 
\textbf{\gls{cer}}~\citep{ceren} from the energy consumption domain, 
\textbf{\gls{climate}}~\citep{de2024graph} and \textbf{\gls{engrad}}~\citep{marisca2024graph} as two multivariate climatic datasets. 
Details on the datasets, data splits and forecasting settings can be found in Appendix~\ref{app:datasets}. 

Regarding the investigated models, we consider three different hybrid global-local architectures (Sec.~\ref{sec:st_models}) distinguished by three different implementations of the propagation layer~(Eq.~\ref{eq:stp}). 
In particular: \textbf{1)} a \gls{rnn} with \glspl{gru} cells~\citep{cho2014learning}~(\textbf{\gls{rnn}}), \textbf{2)} a \gls{stgnn} stacking \gls{gru} and  anisotropic message-passing layers~\citep{bresson2017residual}~(\textbf{\gls{stgnn}}), and \textbf{3)} a \gls{gru} followed by multi-head attention across sequences~\citep{vaswani2017attention}~(\textbf{\gls{statt}}). 
Such hybrid architectures are representative of the current state of the art in the considered benchmarks and of the dominant deep learning frameworks for processing sets of related time series~\citep{benidis2022deep, cini2023graph}. 
Implementation and architectural details are provided in Appendix~\ref{app:models}, while details on the experimental settings can be found in Appendix~\ref{app:details}.

\subsection{Time series forecasting benchmarks}
\label{subsec:benchmarks}
\begin{table}[t]
\caption{Forecasting test error under optimal model size and learning rate (5 runs, $\pm 1std$). Methods equal to or better than the corresponding standard architectures (\acrshort{rnn}, \acrshort{stgnn}, \acrshort{statt}) with embeddings (\textsc{+ Emb.}) are in bold. The best-performing method within each dataset and model is in red. `\textsc{+ Regularization}' denotes the addition of that specific regularization only, on top of `\textsc{+ Emb.}' .}
\label{tab:pred_performance}
\begin{center}
\begin{small}
\setlength{\tabcolsep}{3.5pt}
\setlength{\aboverulesep}{0.5pt}
\setlength{\belowrulesep}{0.5pt}
\begin{sc}
\resizebox{0.99\textwidth}{!}{%
\begin{tabular}{l| c | c | c | c | c | c}
\toprule[1pt]
\multicolumn{1}{c|}{\sc Dataset} & \multicolumn{1}{c|}{\gls{la}} & \multicolumn{1}{c|}{\gls{bay}} & \multicolumn{1}{c|}{\gls{cer}} & \multicolumn{1}{c|}{\gls{air}} & \multicolumn{1}{c|}{\gls{climate}} & \multicolumn{1}{c}{\gls{engrad}} \\
\cmidrule(l{15pt}r{15pt}){1-1} \cmidrule{2-7}
\multicolumn{1}{c|}{\sc Model} & {\acrshort{mae} $\downarrow$} & {\acrshort{mae} $\downarrow$} & {\acrshort{mae} $\downarrow$} & {\acrshort{mae} $\downarrow$} & {\acrshort{mmre} $\downarrow$} & {\acrshort{mmre} $\downarrow$}\\

\arrayrulecolor{black}\midrule[1pt]
\acrshort{rnn}     & $3.556_{\pm .004}$          & $1.774_{\pm .002}$          & $0.4453_{\pm .0010}$          & $13.279_{\pm .042}$          & $19.66_{\pm .01}$          & $31.04_{\pm .04}$ \\
+ Emb.  & $3.148_{\pm .011}$          & $1.593_{\pm .004}$          & $0.4146_{\pm .0025}$          & $13.247_{\pm .044}$          & $19.41_{\pm .01}$          & $31.01_{\pm .10}$ \\
\arrayrulecolor{black!30}\midrule
+ L1    & $3.149_{\pm .007}$          & $\mathbf{1.590}_{\pm .002}$ & $\mathbf{0.4079}_{\pm .0023}$ & $\mathbf{13.147}_{\pm .047}$ & $19.47_{\pm .01}$          & $31.02_{\pm .11}$ \\
+ L2    & $\mathbf{3.146}_{\pm .009}$ & $\mathbf{1.586}_{\pm .005}$ & $\mathbf{0.4058}_{\pm .0011}$ & $\mathbf{13.181}_{\pm .026}$ & $19.42_{\pm .01}$          & $\mathbf{30.97}_{\pm .06}$ \\
+ Clust. & $\mathbf{3.138}_{\pm .015}$ & $\mathbf{1.588}_{\pm .007}$ & $\mathbf{0.4115}_{\pm .0025}$ & $\mathbf{13.231}_{\pm .044}$ & $\mathbf{19.41}_{\pm .01}$ & $31.05_{\pm .07}$ \\
+ Drop. & $\mathbf{3.147}_{\pm .012}$ & ${\color{BrownRed}\mathbf{1.580}}_{\pm .007}$ & $\mathbf{0.4104}_{\pm .0005}$ & $\mathbf{13.114}_{\pm .038}$ & $19.43_{\pm .01}$          & $\mathbf{30.99}_{\pm .09}$ \\
+ Vari. & ${\color{BrownRed}\mathbf{3.132}}_{\pm .006}$ & $\mathbf{1.589}_{\pm .006}$ & $\mathbf{0.4050}_{\pm .0006}$ & ${\color{BrownRed}\mathbf{13.113}}_{\pm .029}$ & ${\color{BrownRed}\mathbf{19.39}}_{\pm .00}$ & $\mathbf{30.98}_{\pm .05}$ \\
+ Forg. & $3.149_{\pm .007}$          & $\mathbf{1.590}_{\pm .007}$ & ${\color{BrownRed}\mathbf{0.4049}}_{\pm .0007}$ & $\mathbf{13.185}_{\pm .022}$ & $19.42_{\pm .02}$          & ${\color{BrownRed}\mathbf{30.92}}_{\pm .02}$ \\
\arrayrulecolor{black}\midrule[1pt]
\acrshort{stgnn}   & $3.239_{\pm .017}$          & $1.660_{\pm .003}$          & $0.4275_{\pm .0006}$          & $11.814_{\pm .051}$          & $19.19_{\pm .03}$          & $28.04_{\pm .08}$ \\
+ Emb.  & $3.027_{\pm .009}$          & $1.593_{\pm .004}$          & $0.4144_{\pm .0032}$          & $11.881_{\pm .053}$          & $18.89_{\pm .04}$          & $27.52_{\pm .09}$ \\
\arrayrulecolor{black!30}\midrule
+ L1    & $3.040_{\pm .016}$          & $\mathbf{1.587}_{\pm .005}$ & $\mathbf{0.4039}_{\pm .0009}$ & $\mathbf{11.789}_{\pm .043}$ & $18.92_{\pm .03}$          & $\mathbf{27.45}_{\pm .12}$ \\
+ L2    & $\mathbf{3.023}_{\pm .009}$ & $\mathbf{1.582}_{\pm .003}$ & $\mathbf{0.4016}_{\pm .0014}$ & $\mathbf{11.795}_{\pm .025}$ & $\mathbf{18.87}_{\pm .04}$ & $\mathbf{27.44}_{\pm .14}$ \\
+ Clust. & $\mathbf{3.025}_{\pm .012}$ & $\mathbf{1.580}_{\pm .005}$ & $\mathbf{0.4075}_{\pm .0020}$ & $\mathbf{11.876}_{\pm .053}$ & $\mathbf{18.85}_{\pm .04}$ & $27.60_{\pm .14}$ \\
+ Drop. & $3.036_{\pm .011}$          & $\mathbf{1.575}_{\pm .006}$ & $\mathbf{0.4042}_{\pm .0008}$ & ${\color{BrownRed}\mathbf{11.712}}_{\pm .016}$ & $18.93_{\pm .04}$          & ${\color{BrownRed}\mathbf{27.41}}_{\pm .06}$ \\
+ Vari. & ${\color{BrownRed}\mathbf{3.013}}_{\pm .005}$ & ${\color{BrownRed}\mathbf{1.566}}_{\pm .003}$ & ${\color{BrownRed}\mathbf{0.3989}}_{\pm .0012}$ & $\mathbf{11.768}_{\pm .026}$ & $\mathbf{18.85}_{\pm .02}$ & $27.53_{\pm .10}$ \\
+ Forg. & $3.050_{\pm .017}$          & $\mathbf{1.578}_{\pm .006}$ & $\mathbf{0.4026}_{\pm .0006}$ & $\mathbf{11.793}_{\pm .040}$ & ${\color{BrownRed}\mathbf{18.83}}_{\pm .02}$ & $\mathbf{27.47}_{\pm .12}$ \\
\arrayrulecolor{black}\midrule[1pt]
\acrshort{statt}  & $3.538_{\pm .004}$          & $1.776_{\pm .002}$          & $0.4479_{\pm .0014}$          & $13.341_{\pm .330}$          & $19.74_{\pm .02}$          & $29.25_{\pm .09}$ \\
+ Emb.  & $3.074_{\pm .020}$          & $1.616_{\pm .012}$          & $0.4143_{\pm .0022}$          & $12.973_{\pm .402}$          & $19.10_{\pm .03}$ & $28.32_{\pm .08}$ \\
\arrayrulecolor{black!30}\midrule
+ L1    & $\mathbf{3.061}_{\pm .021}$ & $\mathbf{1.590}_{\pm .005}$ & $\mathbf{0.4072}_{\pm .0009}$ & $\mathbf{12.799}_{\pm .697}$ & $19.27_{\pm .04}$          & $\mathbf{28.15}_{\pm .33}$ \\
+ L2    & $\mathbf{3.058}_{\pm .017}$ & $\mathbf{1.593}_{\pm .004}$ & $\mathbf{0.4061}_{\pm .0004}$ & $13.542_{\pm .606}$          & $19.17_{\pm .06}$          & $\mathbf{28.12}_{\pm .19}$ \\
+ Clust. & $\mathbf{3.061}_{\pm .021}$ & $\mathbf{1.590}_{\pm .008}$ & $\mathbf{0.4084}_{\pm .0017}$ & $\mathbf{12.535}_{\pm .240}$ & ${\color{BrownRed}\mathbf{19.06}}_{\pm .01}$ & $\mathbf{27.96}_{\pm .17}$ \\
+ Drop. & $\mathbf{3.058}_{\pm .005}$ & ${\color{BrownRed}\mathbf{1.565}}_{\pm .004}$ & $\mathbf{0.4068}_{\pm .0012}$ & ${\color{BrownRed}\mathbf{12.443}}_{\pm .482}$ & $19.30_{\pm .08}$          & $\mathbf{28.12}_{\pm .23}$ \\
+ Vari. & ${\color{BrownRed}\mathbf{3.041}}_{\pm .009}$ & $\mathbf{1.571}_{\pm .005}$ & ${\color{BrownRed}\mathbf{0.4030}}_{\pm .0014}$ & $\mathbf{12.616}_{\pm 1.145}$ & $\mathbf{19.10}_{\pm .06}$ & $\mathbf{28.01}_{\pm .13}$ \\
+ Forg. & $\mathbf{3.058}_{\pm .011}$ & $\mathbf{1.579}_{\pm .003}$ & $\mathbf{0.4062}_{\pm .0012}$ & $\mathbf{12.455}_{\pm .416}$ & $19.16_{\pm .02}$          & ${\color{BrownRed}\mathbf{27.94}}_{\pm .14}$ \\
\arrayrulecolor{black}\bottomrule[1pt]
\end{tabular}
}
\end{sc}
\end{small}
\end{center}             
\end{table}

\vspace{-0.33em}
In our first experiment, we consider the problem of time series forecasting in a transductive setting, i.e., the set of time series to be forecast is the same set observed at training time. 
The forecasting horizon is dataset-dependent and reported in Tab.~\ref{tab:datasets} of Appendix~\ref{app:datasets}. We use models without local parameters and with un-regularized local embeddings as reference baselines. Then, we evaluate performance with regularized local embeddings, adopting the different strategies detailed in Sec.~\ref{subsec:local-regularizations}. For model selection, a hyperparameter search on the hidden size (i.e., the number of units in each model's hidden layer) and learning rate has been carried out independently for each model variant and dataset. All regularization hyperparameters have been set as detailed in Appendix~\ref{app:details}. 

\begin{wraptable}[16]{r}{0.275\linewidth}
    \vspace{-.75em}
    \begin{minipage}{\textwidth}
\captionof{table}{Summary statistics for Tab.~\ref{tab:pred_performance}. Columns indicate the average improvement (\%) over the reference un-regularized model and mean relative rank. Best is in red, second in bold.}
\label{tab:perf_avg}
\begin{center}
\begin{small}
\setlength{\tabcolsep}{3.5pt}
\setlength{\aboverulesep}{0.5pt}
\setlength{\belowrulesep}{0.5pt}
\begin{sc}
\resizebox{\textwidth}{!}{%
\begin{tabular}{l| c | c}
\toprule[1pt]
\multicolumn{1}{c|}{\sc Reg.} & \multicolumn{1}{c|}{\%impr.} & \multicolumn{1}{c}{Rank}\\
\arrayrulecolor{black}\midrule[1pt]
L1          & $0.58$                            & $4.6$             \\
L2          & $0.45$                            & $3.4$             \\
Clust.      & $0.67$                            & $4.0$             \\
Drop.       & $0.94$                            & $3.2$             \\
Vari.       & ${\color{BrownRed}\mathbf{1.21}}$ & ${\color{BrownRed}\mathbf{2.0}}$    \\
Forg.       & $\mathbf{0.95}$                   & $\mathbf{3.0}$ \\
\arrayrulecolor{black}\bottomrule[1pt]
\end{tabular}
}
\end{sc}
\end{small}
\end{center}             
\end{minipage}
\end{wraptable}
Tab.~\ref{tab:pred_performance} shows the obtained results, while Tab.~\ref{tab:perf_avg} provides a concise summary of the performance of each regularization across experimental settings. 
Regardless of the strategy, regularizing the learning of local embeddings provides consistent performance improvements over non-regularized models in most datasets~\ref{F1}. Considering the negligible computational overhead and ease of implementation, these results support the adoption of such techniques as standard practice. While there is no clear winner among the different regularization methods, the \textit{variational} regularization appears to be the most effective, on average; followed by \textit{forgetting} and \textit{dropout}. This suggests that methods perturbing the embedding values can be more effective in preventing co-adaptation~\ref{F2}~\ref{F3}.  
We can speculate that the effectiveness of the \textit{variational} regularization stems from it jointly perturbing embeddings' values while also providing structure to the embedding space. 
Additional results on the impact of regularizing embeddings when Eq.~\ref{eq:stp} is implemented by a simple \gls{mlp} are reported in Appendix~\ref{app:mlp}, while results regarding the combination of multiple regularization techniques can be found in Appendix~\ref{app:combo}. 

\subsection{Sensitivity analysis and learning curves}
\label{subsec:overfit}

\begin{figure}
     \centering
      \includegraphics[width=\textwidth]{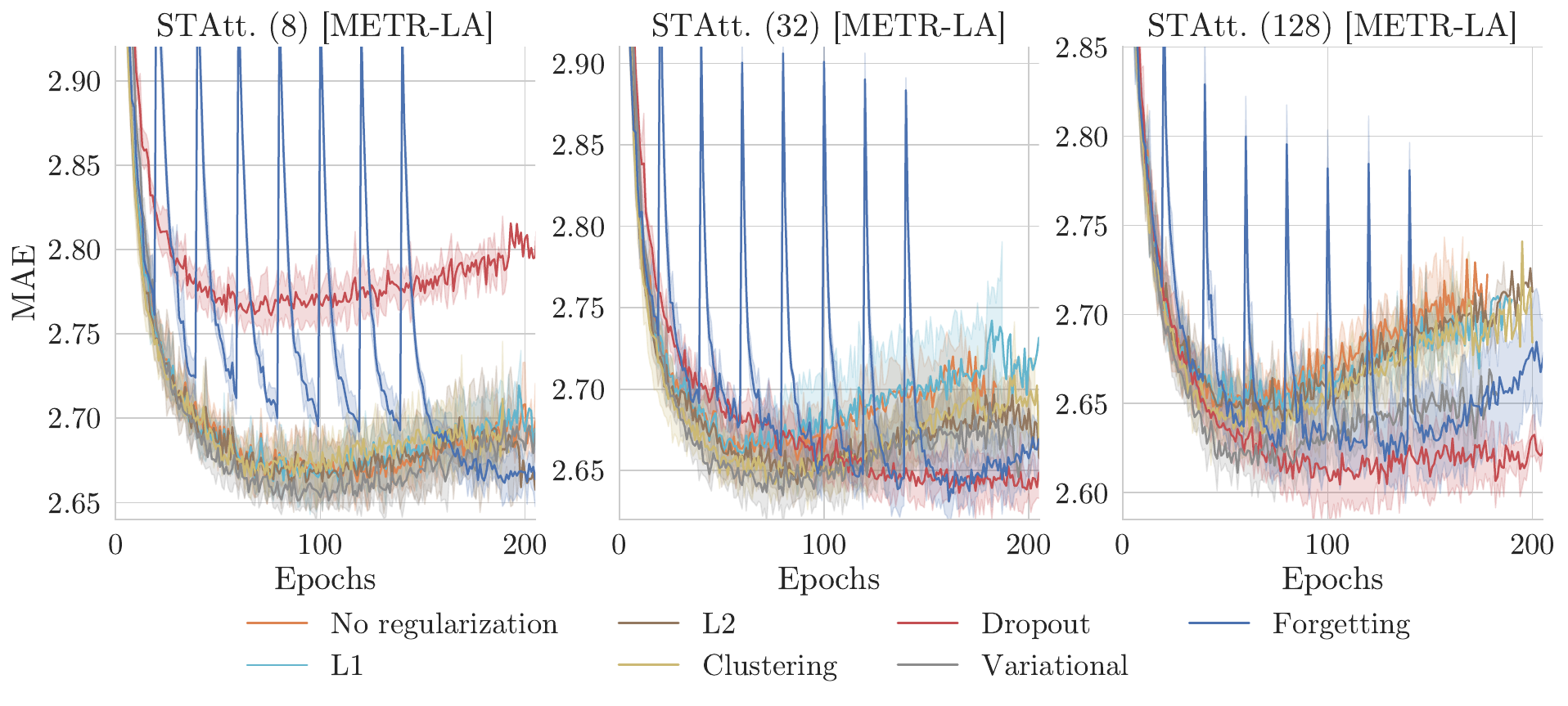}
    \vspace{-.7cm}
    \caption{Validation curves for different training scenarios (5 runs, $\pm 1std$). Plot names follow the convention \textit{model (embedding size) [dataset]}.}
    \label{fig:overfit}
\end{figure}
To provide additional insight, we investigate how different regularizations affect the learning curve across different embedding sizes. In doing so, we fix the shared model hidden size to $d_{h} = 64$ and learning rate to $lr = 0.00075$. Specifically, Fig.~\ref{fig:overfit} shows the validation \gls{mae} across training epochs for some example scenarios; different colors correspond to different regularization strategies. For completeness, additional plots in complementary settings are reported in Appendix~\ref{app:additional_exp}. Comparing the different regularizations, we can see that dropout~(red) and forgetting~(blue) significantly affect the learning curves, even with large embeddings~(Fig.~\ref{fig:overfit}, right). However, dropout, as one could expect, might be problematic when applied to embeddings of limited size~(Fig.~\ref{fig:overfit}, left). The forgetting regularization is robust across the different configurations. The variational regularization appears less disruptive of the learning dynamics. The remaining regularizations (i.e., L1, L2, and clustering) have little impact on the learning curve. 
Overall, regularization strategies that perturb the local embeddings seem more likely to positively affect the learning curves, which might be a reason for the superior performance shown in Tab.~\ref{tab:perf_avg}.

\subsection{Transfer learning}
\label{subsec:transfer}

\begin{table}[t]
\caption{Forecasting test error (\acrshort{mae}) in transfer learning (5 runs, $\pm 1std$, \acrshort{stgnn}). Within each target dataset, different rows pertain to different fine-tuning data lengths. Bold denotes methods equal to or better than the baseline~(\textsc{+ Emb.}), best method is in red. `\textsc{+ Regularization}' denotes the addition of that specific regularization only, on top of `\textsc{+ Emb.}' .}
\label{tab:transfer}
\begin{center}
\begin{small}
\setlength{\tabcolsep}{3.5pt}
\setlength{\aboverulesep}{0.6pt}
\setlength{\belowrulesep}{0.6pt}
\begin{sc}
\resizebox{\textwidth}{!}{%
\begin{tabular}{l| c | c | c | c | c | c | c | c}
\toprule[1pt]
 \multicolumn{1}{c}{} & \multicolumn{1}{c|}{Tr.\ size} & + Emb. & + L1 & + L2 & + Clust. & + Dropout & + Varia. & + Forget.\\

\arrayrulecolor{black}\midrule[1pt]
\multirow{5}{*}{\rotatebox[origin=c]{90}{\gls{pems3}}} 
    & Zero-shot & $26.19_{\pm .98}$ & $\mathbf{26.06}_{\pm .61}$ & $\mathbf{25.26}_{\pm .68}$ & $\mathbf{25.18}_{\pm .40}$ & ${\color{BrownRed}\mathbf{21.45}}_{\pm .26}$ & $\mathbf{22.71}_{\pm .13}$ & $\mathbf{25.08}_{\pm .46}$ \\
    & 1 day     & $18.92_{\pm .08}$ & $\mathbf{18.89}_{\pm .10}$ & $\mathbf{18.87}_{\pm .08}$ & $\mathbf{18.77}_{\pm .05}$ & ${\color{BrownRed}\mathbf{17.99}}_{\pm .02}$ & $\mathbf{18.17}_{\pm .07}$ & $\mathbf{18.55}_{\pm .09}$ \\
    & 3 days    & $18.41_{\pm .04}$ & $\mathbf{18.41}_{\pm .07}$ & $\mathbf{18.35}_{\pm .05}$ & $18.53_{\pm .06}$ & ${\color{BrownRed}\mathbf{17.86}}_{\pm .05}$ & $\mathbf{17.91}_{\pm .03}$ & $\mathbf{18.19}_{\pm .03}$ \\
    & 1 week    & $17.53_{\pm .02}$ & $\mathbf{17.52}_{\pm .05}$ & $\mathbf{17.47}_{\pm .07}$ & $17.59_{\pm .06}$ & $\mathbf{17.31}_{\pm .04}$ & ${\color{BrownRed}\mathbf{17.26}}_{\pm .02}$ & $\mathbf{17.34}_{\pm .03}$ \\
    & 2 weeks   & $17.34_{\pm .03}$ & $\mathbf{17.34}_{\pm .03}$ & $\mathbf{17.28}_{\pm .04}$ & $17.37_{\pm .04}$ & $\mathbf{17.20}_{\pm .02}$ & ${\color{BrownRed}\mathbf{17.16}}_{\pm .02}$ & $\mathbf{17.20}_{\pm .05}$ \\
\arrayrulecolor{black}\midrule
\multirow{5}{*}{\rotatebox[origin=c]{90}{\gls{pems4}}} 
    & Zero-shot & $29.23_{\pm .43}$ & $\mathbf{29.09}_{\pm .26}$ & $29.26_{\pm .47}$ & $30.03_{\pm .26}$ & ${\color{BrownRed}\mathbf{26.13}}_{\pm .31}$ & $\mathbf{27.56}_{\pm .31}$ & $\mathbf{28.44}_{\pm .34}$ \\
    & 1 day     & $23.93_{\pm .16}$ & $24.00_{\pm .10}$ & $23.96_{\pm .16}$ & $\mathbf{23.71}_{\pm .05}$ & ${\color{BrownRed}\mathbf{23.04}}_{\pm .04}$ & $\mathbf{23.33}_{\pm .12}$ & $\mathbf{23.54}_{\pm .13}$ \\
    & 3 days    & $23.18_{\pm .12}$ & $23.22_{\pm .14}$ & $23.27_{\pm .10}$ & $\mathbf{22.97}_{\pm .06}$ & ${\color{BrownRed}\mathbf{22.67}}_{\pm .03}$ & $\mathbf{22.87}_{\pm .06}$ & $\mathbf{22.86}_{\pm .08}$ \\
    & 1 week    & $22.47_{\pm .07}$ & $\mathbf{22.46}_{\pm .05}$ & $\mathbf{22.45}_{\pm .07}$ & $22.49_{\pm .07}$ & $\mathbf{22.29}_{\pm .03}$ & $\mathbf{22.38}_{\pm .05}$ & ${\color{BrownRed}\mathbf{22.26}}_{\pm .03}$ \\
    & 2 weeks   & $21.92_{\pm .04}$ & $21.93_{\pm .04}$ & $21.93_{\pm .05}$ & $22.00_{\pm .01}$ & $21.96_{\pm .03}$ & $22.04_{\pm .04}$ & ${\color{BrownRed}\mathbf{21.79}}_{\pm .03}$ \\
\arrayrulecolor{black}\midrule
\multirow{5}{*}{\rotatebox[origin=c]{90}{\gls{pems7}}} 
    & Zero-shot & $57.40_{\pm 6.70}$ & $\mathbf{55.30}_{\pm 3.97}$ & $\mathbf{53.97}_{\pm 3.37}$ & $\mathbf{56.92}_{\pm 1.62}$ & ${\color{BrownRed}\mathbf{34.19}}_{\pm .78}$ & $\mathbf{42.99}_{\pm 2.83}$ & $\mathbf{42.67}_{\pm 2.64}$ \\
    & 1 day     & $29.61_{\pm .17}$ & $29.63_{\pm .24}$ & $\mathbf{29.37}_{\pm .27}$ & $\mathbf{29.00}_{\pm .32}$ & ${\color{BrownRed}\mathbf{27.25}}_{\pm .12}$ & $\mathbf{28.30}_{\pm .41}$ & $\mathbf{28.38}_{\pm .19}$ \\
    & 3 days    & $27.65_{\pm .14}$ & $27.74_{\pm .14}$ & $\mathbf{27.56}_{\pm .14}$ & $\mathbf{27.56}_{\pm .14}$ & ${\color{BrownRed}\mathbf{26.37}}_{\pm .10}$ & $\mathbf{26.92}_{\pm .21}$ & $\mathbf{26.92}_{\pm .08}$ \\
    & 1 week    & $26.60_{\pm .04}$ & $26.66_{\pm .09}$ & $\mathbf{26.60}_{\pm .07}$ & $26.90_{\pm .14}$ & ${\color{BrownRed}\mathbf{25.87}}_{\pm .13}$ & $\mathbf{26.20}_{\pm .17}$ & $\mathbf{26.03}_{\pm .06}$ \\
    & 2 weeks   & $25.84_{\pm .04}$ & $25.90_{\pm .07}$ & $\mathbf{25.76}_{\pm .04}$ & $26.31_{\pm .24}$ & $\mathbf{25.53}_{\pm .15}$ & $\mathbf{25.66}_{\pm .17}$ & ${\color{BrownRed}\mathbf{25.41}}_{\pm .04}$ \\
\arrayrulecolor{black}\midrule
\multirow{5}{*}{\rotatebox[origin=c]{90}{\gls{pems8}}} 
    & Zero-shot & $25.41_{\pm .62}$ & $\mathbf{25.26}_{\pm 1.09}$ & $25.90_{\pm .47}$ & $25.60_{\pm .44}$ & ${\color{BrownRed}\mathbf{21.04}}_{\pm .17}$ & $\mathbf{22.60}_{\pm .37}$ & $\mathbf{23.90}_{\pm .43}$ \\
    & 1 day     & $18.91_{\pm .09}$ & $\mathbf{18.88}_{\pm .05}$ & $\mathbf{18.85}_{\pm .08}$ & $\mathbf{18.64}_{\pm .08}$ & ${\color{BrownRed}\mathbf{17.96}}_{\pm .06}$ & $\mathbf{18.23}_{\pm .12}$ & $\mathbf{18.40}_{\pm .13}$ \\
    & 3 days    & $18.11_{\pm .10}$ & $18.16_{\pm .13}$ & $18.15_{\pm .08}$ & $\mathbf{18.03}_{\pm .04}$ & ${\color{BrownRed}\mathbf{17.55}}_{\pm .06}$ & $\mathbf{17.77}_{\pm .07}$ & $\mathbf{17.83}_{\pm .14}$ \\
    & 1 week    & $17.33_{\pm .08}$ & $\mathbf{17.28}_{\pm .06}$ & $\mathbf{17.33}_{\pm .03}$ & $17.40_{\pm .04}$ & $\mathbf{17.22}_{\pm .06}$ & $\mathbf{17.31}_{\pm .07}$ & ${\color{BrownRed}\mathbf{17.11}}_{\pm .06}$ \\
    & 2 weeks   & $17.13_{\pm .06}$ & $17.18_{\pm .14}$ & $17.15_{\pm .07}$ & $17.25_{\pm .03}$ & $17.15_{\pm .08}$ & $17.20_{\pm .09}$ & ${\color{BrownRed}\mathbf{16.99}}_{\pm .05}$ \\
\arrayrulecolor{black}\bottomrule[1pt]
\end{tabular}
}
\end{sc}
\end{small}
\end{center}             
\end{table}
Our second main experiment consists of time series forecasting under a transfer learning setting, adapted from previous works~\citep{cini2023taming}. This benchmark aims to verify the impact of regularizing the local embeddings on the transferability of the global (shared) processing blocks. We consider the $4$ PEMS benchmark datasets from \citet{guo2021learning} (i.e., \gls{pems3}, \gls{pems4}, \gls{pems7}, \gls{pems8}) and a reference \gls{stgnn} model with local embeddings. For each subset of three datasets, we first train the entire model, then we reset the embeddings to their initial values and fine-tune them exclusively on the held-out dataset. During fine-tuning, shared parameters are kept frozen. The different regularizations are applied only during the initial training procedure and disabled during fine-tuning. Tab.~\ref{tab:transfer} reports the results for the un-regularized and regularized models as the amount of data for fine-tuning varies~(zero-shot refers to no fine-tuning at all). It is evident how dropout, variational regularization, and forgetting are the most reliable and effective strategies~\ref{F2}. 
Notably, dropout excels in the zero-shot setting, while forgetting is consistent across fine-tuning lengths and excels when more data are available~\ref{F3}. This suggests dropout can be particularly useful when data from the transfer domain is scarce or absent. 
Conversely, regularizations based on structuring the embedding space, i.e., L1, L2, and the \textit{clustering} regularization, show mixed results compared to the unconstrained model.  

\subsection{Robustness to local parameter perturbation}
\label{subsec:perturbation}

\begin{figure}[t]
    \centering
    \includegraphics[width=\textwidth]{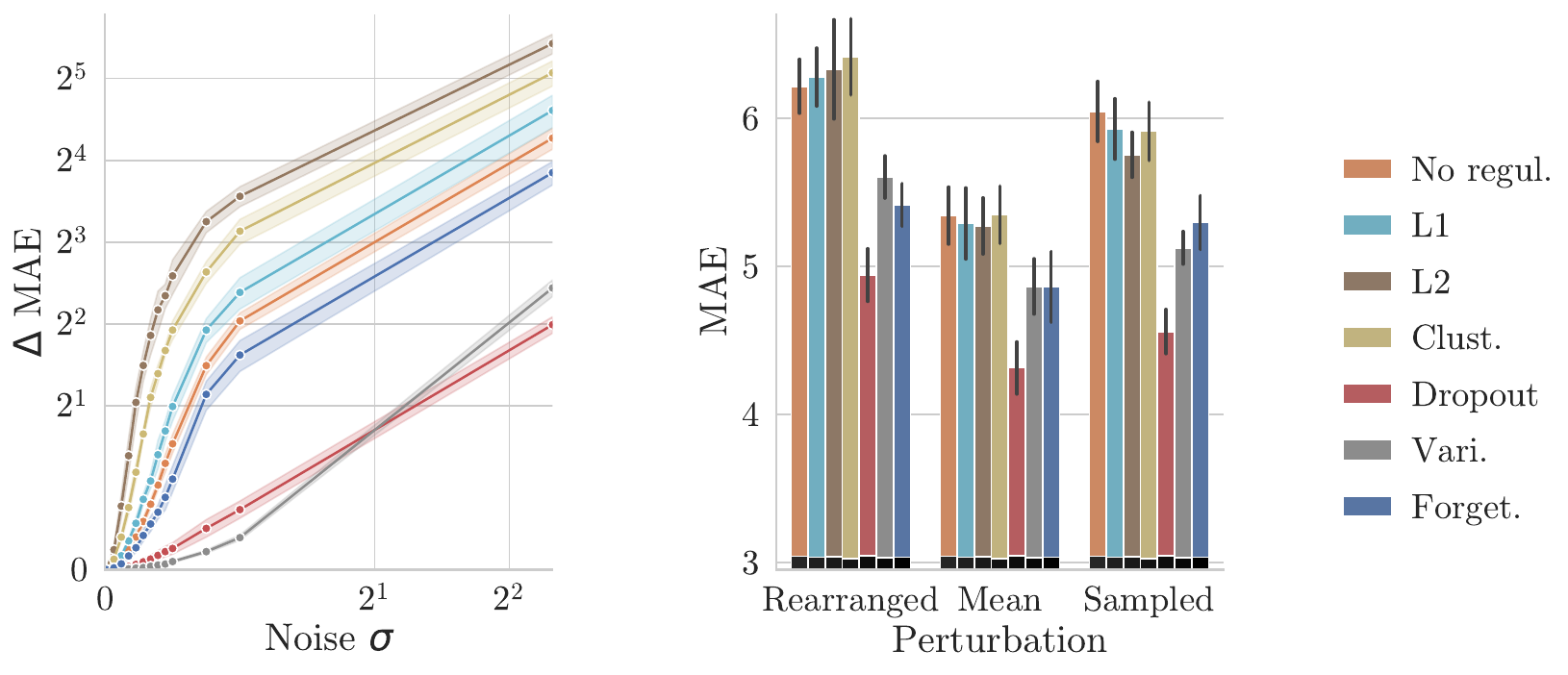}
    \vspace{-.35cm}
    \caption{Test performance degradation on embeddings perturbation (5 runs, $\pm 1std$, \gls{stgnn}, \gls{la}). \textbf{Left:} Adding zero-mean Gaussian noise. \textbf{Right:} (\textbf{Left}) random shuffling, (\textbf{Middle}) replaced with their mean, and (\textbf{Right}) replaced by a draw from their sample normal.}
    \label{fig:perturbation}
\end{figure}
Finally, we investigate how a model's forecasting accuracy is affected by the perturbation of the learned local embeddings. This provides insights into the robustness of the shared learned parameters and can serve as a proxy for the effectiveness of different regularizations in preventing co-adaptation. 
To avoid penalizing regularizations that are sensitive to the weights' magnitude ~(i.e., L2, L1), we consider perturbations that do not impact the scale of the learned representations. In particular, we experiment with four such strategies: adding zero-mean Gaussian noise to the embeddings (\textit{Noise} $\sigma$), randomly shuffling embeddings across sequences (\textit{Rearranged}), replacing each embedding with their sample mean (\textit{Mean}), resampling each embedding from their sample normal distribution (\textit{Sampled}) (see Appendix~\ref{app:perturbation} for a formal definition). Fig.~\ref{fig:perturbation} shows the results of the analysis after training an \gls{stgnn} on the \gls{la} dataset (complementary settings are illustrated in Appendix~\ref{app:additional_exp}). As one might expect, regularization methods that actively perturb the embeddings while learning (i.e., dropout, forgetting, variational) consistently result in more robust models under all the considered perturbations~\ref{F2}~\ref{F3}. Conversely, other regularizations have marginal, or even negative, impact. We observe some consistency between the results observed in Fig.~\ref{fig:perturbation} and performance in transfer learning, shown in Tab.~\ref{tab:transfer}. 

\subsection{Discussion}
Considering the observed empirical results, we can summarize a set of practical recommendations that can provide practitioners with a starting point for model search. In transductive settings, \textit{variational} regularization performs reliably across benchmarks and it is a safe choice. Moreover, we suggest favoring techniques that perturb the embedding values during training, e.g., \textit{variational} regularization, \textit{forgetting} or \textit{dropout}. In the context of transfer learning, our results suggest the adoption of \textit{dropout}, when fine-tuning the model on a low amount of data, while \textit{forgetting} or \textit{variational} regularization can give better results as more data become available. The recommendation of prioritizing techniques that perturb the embeddings at training time remains valid in the transfer learning settings. 

\section{Conclusions}\label{sec:conclusions}
This paper highlights the importance of regularizing the learning of local embeddings in modern deep-learning architectures for time series forecasting. Our empirical study, across diverse datasets and scenarios, provides clear evidence that this practice is beneficial for a variety of reference architectures, representative of the state of the art. Notably, even simple techniques, when applied to local embeddings, can yield consistent performance gains. Furthermore, we observed that methods that actively perturb the embeddings at training time, such as dropout, \textit{variational} regularization, and \textit{forgetting}, consistently rank among the top performers, both in transductive and transfer learning settings. The use of this property as a guideline in designing new regularization methods is further supported by the effectiveness of the unprecedentedly considered \textit{forgetting} regularization. 
Overall, our findings offer a strong empirical argument for the adoption of embedding regularization as standard practice when designing hybrid global-local architectures for time series. We believe this is an important building block towards developing more robust and transferable models for spatiotemporal data and designing foundation models for related time series processing.

\paragraph{Limitations and future works} 
While it is clear that these regularizations provide consistent advantages, it is difficult to strongly identify the best-performing method in the different scenarios. 
Hence, future works might focus on finding strategies that combine the best qualities of the different methods.
Furthermore, future studies could try to provide methods to quantitatively and analytically characterize the co-adaptation of global and local parameters, for which we can only identify indirect signs. 
In particular, they might design experiments to detect the degeneration of local components into identifiers. 
Moreover, future research could explore how the efficacy of different regularization methods varies with the number of input time series and different downstream tasks. 

\subsubsection*{Acknowledgments}
Partly supported by International Partnership Program of the Chinese Academy of Sciences under Grant 104GJHZ2022013GC

\bibliography{references}

\begin{thebibliography}{82}
\providecommand{\natexlab}[1]{#1}
\providecommand{\url}[1]{\texttt{#1}}
\expandafter\ifx\csname urlstyle\endcsname\relax
  \providecommand{\doi}[1]{doi: #1}\else
  \providecommand{\doi}{doi: \begingroup \urlstyle{rm}\Url}\fi

\bibitem[Abuduweili et~al.(2021)Abuduweili, Li, Shi, Xu, and Dou]{abuduweili2021adaptive}
Abulikemu Abuduweili, Xingjian Li, Humphrey Shi, Cheng-Zhong Xu, and Dejing Dou.
\newblock Adaptive consistency regularization for semi-supervised transfer learning.
\newblock In \emph{Proceedings of the IEEE/CVF conference on computer vision and pattern recognition}, pp.\  6923--6932, 2021.

\bibitem[Ba et~al.(2016)Ba, Kiros, and Hinton]{ba2016layer}
Jimmy~Lei Ba, Jamie~Ryan Kiros, and Geoffrey~E Hinton.
\newblock Layer normalization.
\newblock \emph{arXiv preprint arXiv:1607.06450}, 2016.

\bibitem[Bai et~al.(2020)Bai, Yao, Li, Wang, and Wang]{bai2020adaptive}
Lei Bai, Lina Yao, Can Li, Xianzhi Wang, and Can Wang.
\newblock Adaptive graph convolutional recurrent network for traffic forecasting.
\newblock \emph{Advances in neural information processing systems}, 33:\penalty0 17804--17815, 2020.

\bibitem[Bai et~al.(2018)Bai, Kolter, and Koltun]{bai2018empirical}
Shaojie Bai, J~Zico Kolter, and Vladlen Koltun.
\newblock An empirical evaluation of generic convolutional and recurrent networks for sequence modeling.
\newblock \emph{arXiv preprint arXiv:1803.01271}, 2018.

\bibitem[Baldock et~al.(2021)Baldock, Maennel, and Neyshabur]{baldock2021deep}
Robert Baldock, Hartmut Maennel, and Behnam Neyshabur.
\newblock Deep learning through the lens of example difficulty.
\newblock \emph{Advances in Neural Information Processing Systems}, 34:\penalty0 10876--10889, 2021.

\bibitem[Benidis et~al.(2022)Benidis, Rangapuram, Flunkert, Wang, Maddix, Turkmen, Gasthaus, Bohlke-Schneider, Salinas, Stella, et~al.]{benidis2022deep}
Konstantinos Benidis, Syama~Sundar Rangapuram, Valentin Flunkert, Yuyang Wang, Danielle Maddix, Caner Turkmen, Jan Gasthaus, Michael Bohlke-Schneider, David Salinas, Lorenzo Stella, et~al.
\newblock Deep learning for time series forecasting: Tutorial and literature survey.
\newblock \emph{ACM Computing Surveys}, 55\penalty0 (6):\penalty0 1--36, 2022.

\bibitem[Biewald(2020)]{wandb}
Lukas Biewald.
\newblock Experiment tracking with weights and biases, 2020.
\newblock URL \url{https://www.wandb.com/}.
\newblock Software available from wandb.com.

\bibitem[Bresson \& Laurent(2017)Bresson and Laurent]{bresson2017residual}
Xavier Bresson and Thomas Laurent.
\newblock Residual gated graph convnets.
\newblock \emph{arXiv preprint arXiv:1711.07553}, 2017.

\bibitem[Cao et~al.(2018)Cao, Wang, Li, Zhou, Li, and Li]{cao2018brits}
Wei Cao, Dong Wang, Jian Li, Hao Zhou, Lei Li, and Yitan Li.
\newblock Brits: Bidirectional recurrent imputation for time series.
\newblock \emph{Advances in neural information processing systems}, 31, 2018.

\bibitem[{CER}(2016)]{ceren}
Commission for {E}nergy~{R}egulation. {CER}.
\newblock {CER} {S}mart {M}etering {P}roject - {E}lectricity {C}ustomer {B}ehaviour {T}rial, 2009-2010 [dataset].
\newblock \emph{Irish Social Science Data Archive. SN: 0012-00}, 2016.
\newblock URL \url{https://www.ucd.ie/issda/data/commissionforenergyregulationcer/}.

\bibitem[Chen et~al.(2023)Chen, Long, Jiang, Liu, and Zhang]{chen2023foundation}
Shengchao Chen, Guodong Long, Jing Jiang, Dikai Liu, and Chengqi Zhang.
\newblock Foundation models for weather and climate data understanding: A comprehensive survey.
\newblock \emph{arXiv preprint arXiv:2312.03014}, 2023.

\bibitem[Cho et~al.(2014{\natexlab{a}})Cho, Van~Merri{\"e}nboer, Bahdanau, and Bengio]{cho2014properties}
Kyunghyun Cho, Bart Van~Merri{\"e}nboer, Dzmitry Bahdanau, and Yoshua Bengio.
\newblock On the properties of neural machine translation: Encoder-decoder approaches.
\newblock \emph{Proceedings of SSST-8, Eighth Workshop on Syntax, Semantics and Structure in Statistical Translation}, 2014{\natexlab{a}}.

\bibitem[Cho et~al.(2014{\natexlab{b}})Cho, van Merri{\"e}nboer, Gul{\c{c}}ehre, Bahdanau, Bougares, Schwenk, and Bengio]{cho2014learning}
Kyunghyun Cho, Bart van Merri{\"e}nboer, {\c{C}}a{\u{g}}lar Gul{\c{c}}ehre, Dzmitry Bahdanau, Fethi Bougares, Holger Schwenk, and Yoshua Bengio.
\newblock Learning phrase representations using rnn encoder--decoder for statistical machine translation.
\newblock In \emph{Proceedings of the 2014 Conference on Empirical Methods in Natural Language Processing (EMNLP)}, pp.\  1724--1734, 2014{\natexlab{b}}.

\bibitem[Cini \& Marisca(2022)Cini and Marisca]{tsl}
Andrea Cini and Ivan Marisca.
\newblock {Torch Spatiotemporal}, 3 2022.
\newblock URL \url{https://github.com/TorchSpatiotemporal/tsl}.

\bibitem[Cini et~al.(2022)Cini, Marisca, and Alippi]{cini2022filling}
Andrea Cini, Ivan Marisca, and Cesare Alippi.
\newblock Filling the g\_ap\_s: multivariate time series imputation by graph neural networks.
\newblock In \emph{International Conference on Learning Representations}, 2022.

\bibitem[Cini et~al.(2023{\natexlab{a}})Cini, Marisca, Zambon, and Alippi]{cini2023graph}
Andrea Cini, Ivan Marisca, Daniele Zambon, and Cesare Alippi.
\newblock Graph deep learning for time series forecasting.
\newblock \emph{arXiv preprint arXiv:2310.15978}, 2023{\natexlab{a}}.

\bibitem[Cini et~al.(2023{\natexlab{b}})Cini, Marisca, Zambon, and Alippi]{cini2023taming}
Andrea Cini, Ivan Marisca, Daniele Zambon, and Cesare Alippi.
\newblock Taming local effects in graph-based spatiotemporal forecasting.
\newblock \emph{Advances in Neural Information Processing Systems}, 36, 2023{\natexlab{b}}.

\bibitem[Clevert et~al.(2016)Clevert, Unterthiner, and Hochreiter]{clevert2016elu}
Djork{-}Arn{\'{e}} Clevert, Thomas Unterthiner, and Sepp Hochreiter.
\newblock Fast and accurate deep network learning by exponential linear units (elus).
\newblock In Yoshua Bengio and Yann LeCun (eds.), \emph{4th International Conference on Learning Representations, {ICLR} 2016, San Juan, Puerto Rico, May 2-4, 2016, Conference Track Proceedings}, 2016.
\newblock URL \url{http://arxiv.org/abs/1511.07289}.

\bibitem[De~Felice et~al.(2024)De~Felice, Cini, Zambon, Gusev, and Alippi]{de2024graph}
Giovanni De~Felice, Andrea Cini, Daniele Zambon, Vladimir~V Gusev, and Cesare Alippi.
\newblock Graph-based virtual sensing from sparse and partial multivariate observations.
\newblock In \emph{International Conference on Learning Representations}, 2024.

\bibitem[Deihim et~al.(2023)Deihim, Alonso, and Apostolopoulou]{deihim2023sttre}
Azad Deihim, Eduardo Alonso, and Dimitra Apostolopoulou.
\newblock Sttre: A spatio-temporal transformer with relative embeddings for multivariate time series forecasting.
\newblock \emph{Neural Networks}, 168:\penalty0 549--559, 2023.

\bibitem[Devlin et~al.(2019)Devlin, Chang, Lee, and Toutanova]{devlin2019bert}
Jacob Devlin, Ming-Wei Chang, Kenton Lee, and Kristina Toutanova.
\newblock Bert: Pre-training of deep bidirectional transformers for language understanding.
\newblock \emph{Proceedings of the 2019 Conference of the North {A}merican Chapter of the Association for Computational Linguistics: Human Language Technologies, Volume 1 (Long and Short Papers)}, 2019.

\bibitem[Dieng et~al.(2018)Dieng, Ranganath, Altosaar, and Blei]{dieng2018noisin}
Adji~Bousso Dieng, Rajesh Ranganath, Jaan Altosaar, and David Blei.
\newblock Noisin: Unbiased regularization for recurrent neural networks.
\newblock In \emph{International Conference on Machine Learning}, pp.\  1252--1261. PMLR, 2018.

\bibitem[Dimoulkas et~al.(2019)Dimoulkas, Mazidi, and Herre]{dimoulkas2019neural}
Ilias Dimoulkas, Peyman Mazidi, and Lars Herre.
\newblock Neural networks for gefcom2017 probabilistic load forecasting.
\newblock \emph{International Journal of Forecasting}, 35\penalty0 (4):\penalty0 1409--1423, 2019.

\bibitem[Falcon \& {The PyTorch Lightning team}(2019)Falcon and {The PyTorch Lightning team}]{Falcon_PyTorch_Lightning_2019}
William Falcon and {The PyTorch Lightning team}.
\newblock {PyTorch Lightning}, March 2019.
\newblock URL \url{https://github.com/Lightning-AI/lightning}.

\bibitem[Gal \& Ghahramani(2016)Gal and Ghahramani]{gal2016theoretically}
Yarin Gal and Zoubin Ghahramani.
\newblock A theoretically grounded application of dropout in recurrent neural networks.
\newblock \emph{Advances in neural information processing systems}, 29, 2016.

\bibitem[Garza \& Mergenthaler-Canseco(2023)Garza and Mergenthaler-Canseco]{garza2023timegpt}
Azul Garza and Max Mergenthaler-Canseco.
\newblock {TimeGPT-1}.
\newblock \emph{arXiv preprint arXiv:2310.03589}, 2023.

\bibitem[Gasparin et~al.(2022)Gasparin, Lukovic, and Alippi]{gasparin2022deep}
Alberto Gasparin, Slobodan Lukovic, and Cesare Alippi.
\newblock Deep learning for time series forecasting: The electric load case.
\newblock \emph{CAAI Transactions on Intelligence Technology}, 7\penalty0 (1):\penalty0 1--25, 2022.

\bibitem[Geirhos et~al.(2020)Geirhos, Jacobsen, Michaelis, Zemel, Brendel, Bethge, and Wichmann]{geirhos2020shortcut}
Robert Geirhos, J{\"o}rn-Henrik Jacobsen, Claudio Michaelis, Richard Zemel, Wieland Brendel, Matthias Bethge, and Felix~A Wichmann.
\newblock Shortcut learning in deep neural networks.
\newblock \emph{Nature Machine Intelligence}, 2\penalty0 (11):\penalty0 665--673, 2020.

\bibitem[Gilmer et~al.(2017)Gilmer, Schoenholz, Riley, Vinyals, and Dahl]{gilmer2017neural}
Justin Gilmer, Samuel~S Schoenholz, Patrick~F Riley, Oriol Vinyals, and George~E Dahl.
\newblock {Neural message passing for quantum chemistry}.
\newblock In \emph{International Conference on Machine Learning}, pp.\  1263--1272. PMLR, 2017.

\bibitem[Grigsby et~al.(2021)Grigsby, Wang, and Qi]{grigsby2021longrange}
Jake Grigsby, Zhe Wang, and Yanjun Qi.
\newblock {Long-Range Transformers for Dynamic Spatiotemporal Forecasting}, 2021.

\bibitem[Guo et~al.(2021)Guo, Lin, Wan, Li, and Cong]{guo2021learning}
Shengnan Guo, Youfang Lin, Huaiyu Wan, Xiucheng Li, and Gao Cong.
\newblock Learning dynamics and heterogeneity of spatial-temporal graph data for traffic forecasting.
\newblock \emph{IEEE Transactions on Knowledge and Data Engineering}, 34\penalty0 (11):\penalty0 5415--5428, 2021.

\bibitem[Hochreiter \& Schmidhuber(1997)Hochreiter and Schmidhuber]{hochreiter1997long}
Sepp Hochreiter and J{\"u}rgen Schmidhuber.
\newblock Long short-term memory.
\newblock \emph{Neural computation}, 9\penalty0 (8):\penalty0 1735--1780, 1997.

\bibitem[Januschowski et~al.(2020)Januschowski, Gasthaus, Wang, Salinas, Flunkert, Bohlke-Schneider, and Callot]{januschowski2020criteria}
Tim Januschowski, Jan Gasthaus, Yuyang Wang, David Salinas, Valentin Flunkert, Michael Bohlke-Schneider, and Laurent Callot.
\newblock Criteria for classifying forecasting methods.
\newblock \emph{International Journal of Forecasting}, 36\penalty0 (1):\penalty0 167--177, 2020.

\bibitem[Jarrett et~al.(2021)Jarrett, Yoon, Bica, Qian, Ercole, and van~der Schaar]{jarrett2021clairvoyance}
Daniel Jarrett, Jinsung Yoon, Ioana Bica, Zhaozhi Qian, Ari Ercole, and Mihaela van~der Schaar.
\newblock Clairvoyance: A pipeline toolkit for medical time series.
\newblock In \emph{International Conference on Learning Representations}, 2021.
\newblock URL \url{https://openreview.net/forum?id=xnC8YwKUE3k}.

\bibitem[Jin et~al.(2023)Jin, Koh, Wen, Zambon, Alippi, Webb, King, and Pan]{jin2023survey}
Ming Jin, Huan~Yee Koh, Qingsong Wen, Daniele Zambon, Cesare Alippi, Geoffrey~I Webb, Irwin King, and Shirui Pan.
\newblock A survey on graph neural networks for time series: Forecasting, classification, imputation, and anomaly detection.
\newblock \emph{arXiv preprint arXiv:2307.03759}, 2023.

\bibitem[Kingma \& Ba(2015)Kingma and Ba]{kingma2015adam}
Diederik Kingma and Jimmy Ba.
\newblock Adam: A method for stochastic optimization.
\newblock In \emph{International Conference on Learning Representations}, 2015.

\bibitem[Krogh \& Hertz(1991)Krogh and Hertz]{krogh1991simple}
Anders Krogh and John Hertz.
\newblock A simple weight decay can improve generalization.
\newblock \emph{Advances in neural information processing systems}, 4, 1991.

\bibitem[LeCun \& Bengio(1998)LeCun and Bengio]{lecun1998convolutional}
Yann LeCun and Yoshua Bengio.
\newblock \emph{{Convolutional Networks for Images, Speech, and Time Series}}, pp.\  255–258.
\newblock MIT Press, Cambridge, MA, USA, 1998.
\newblock ISBN 0262511029.

\bibitem[Li et~al.(2018)Li, Yu, Shahabi, and Liu]{li2018diffusion}
Yaguang Li, Rose Yu, Cyrus Shahabi, and Yan Liu.
\newblock Diffusion convolutional recurrent neural network: Data-driven traffic forecasting.
\newblock In \emph{International Conference on Learning Representations}, 2018.

\bibitem[Liang et~al.(2024)Liang, Wen, Nie, Jiang, Jin, Song, Pan, and Wen]{liang2024foundation}
Yuxuan Liang, Haomin Wen, Yuqi Nie, Yushan Jiang, Ming Jin, Dongjin Song, Shirui Pan, and Qingsong Wen.
\newblock {Foundation models for time series analysis: A tutorial and survey}.
\newblock \emph{arXiv preprint arXiv:2403.14735}, 2024.

\bibitem[Liu et~al.(2023)Liu, Dong, Jiang, Deng, Deng, Chen, and Song]{liu2023spatio}
Hangchen Liu, Zheng Dong, Renhe Jiang, Jiewen Deng, Jinliang Deng, Quanjun Chen, and Xuan Song.
\newblock Spatio-temporal adaptive embedding makes vanilla transformer sota for traffic forecasting.
\newblock In \emph{Proceedings of the 32nd ACM international conference on information and knowledge management}, pp.\  4125--4129, 2023.

\bibitem[Ma et~al.(2023)Ma, Xie, Teng, Wang, Ji, Zhang, and Li]{ma2023histgnn}
Minbo Ma, Peng Xie, Fei Teng, Bin Wang, Shenggong Ji, Junbo Zhang, and Tianrui Li.
\newblock Histgnn: Hierarchical spatio-temporal graph neural network for weather forecasting.
\newblock \emph{Information Sciences}, 648:\penalty0 119580, 2023.

\bibitem[Makridakis et~al.(2020)Makridakis, Spiliotis, and Assimakopoulos]{makridakis2020m4}
Spyros Makridakis, Evangelos Spiliotis, and Vassilios Assimakopoulos.
\newblock {The M4 Competition: 100,000 time series and 61 forecasting methods}.
\newblock \emph{International Journal of Forecasting}, 36\penalty0 (1):\penalty0 54--74, 2020.

\bibitem[Marisca et~al.(2022)Marisca, Cini, and Alippi]{marisca2022learning}
Ivan Marisca, Andrea Cini, and Cesare Alippi.
\newblock Learning to reconstruct missing data from spatiotemporal graphs with sparse observations.
\newblock \emph{Advances in Neural Information Processing Systems}, 35:\penalty0 32069--32082, 2022.

\bibitem[Marisca et~al.(2024)Marisca, Alippi, and Bianchi]{marisca2024graph}
Ivan Marisca, Cesare Alippi, and Filippo~Maria Bianchi.
\newblock Graph-based forecasting with missing data through spatiotemporal downsampling.
\newblock \emph{arXiv preprint arXiv:2402.10634}, 2024.

\bibitem[Mikolov et~al.(2013)Mikolov, Sutskever, Chen, Corrado, and Dean]{mikolov2013distributed}
Tomas Mikolov, Ilya Sutskever, Kai Chen, Greg~S Corrado, and Jeff Dean.
\newblock Distributed representations of words and phrases and their compositionality.
\newblock \emph{Advances in neural information processing systems}, 26, 2013.

\bibitem[Montero-Manso \& Hyndman(2021)Montero-Manso and Hyndman]{montero2021principles}
Pablo Montero-Manso and Rob~J Hyndman.
\newblock Principles and algorithms for forecasting groups of time series: Locality and globality.
\newblock \emph{International Journal of Forecasting}, 37\penalty0 (4):\penalty0 1632--1653, 2021.

\bibitem[Paszke et~al.(2019)Paszke, Gross, Massa, Lerer, Bradbury, Chanan, Killeen, Lin, Gimelshein, Antiga, et~al.]{paszke2019pytorch}
Adam Paszke, Sam Gross, Francisco Massa, Adam Lerer, James Bradbury, Gregory Chanan, Trevor Killeen, Zeming Lin, Natalia Gimelshein, Luca Antiga, et~al.
\newblock Pytorch: An imperative style, high-performance deep learning library.
\newblock \emph{Advances in neural information processing systems}, 32, 2019.

\bibitem[Peng et~al.(2015)Peng, Mou, Li, Chen, Lu, and Jin]{peng2015comparative}
Hao Peng, Lili Mou, Ge~Li, Yunchuan Chen, Yangyang Lu, and Zhi Jin.
\newblock A comparative study on regularization strategies for embedding-based neural networks.
\newblock \emph{Conference on Empirical Methods in Natural Language Processing}, 2015.

\bibitem[Prabowo et~al.(2024)Prabowo, Xue, Shao, Koniusz, and Salim]{prabowo2024traffic}
Arian Prabowo, Hao Xue, Wei Shao, Piotr Koniusz, and Flora~D Salim.
\newblock Traffic forecasting on new roads using spatial contrastive pre-training (scpt).
\newblock \emph{Data Mining and Knowledge Discovery}, 38\penalty0 (3):\penalty0 913--937, 2024.

\bibitem[Salimans \& Kingma(2016)Salimans and Kingma]{salimans2016weight}
Tim Salimans and Durk~P Kingma.
\newblock Weight normalization: A simple reparameterization to accelerate training of deep neural networks.
\newblock \emph{Advances in neural information processing systems}, 29, 2016.

\bibitem[Salinas et~al.(2020)Salinas, Flunkert, Gasthaus, and Januschowski]{salinas2020deepar}
David Salinas, Valentin Flunkert, Jan Gasthaus, and Tim Januschowski.
\newblock Deepar: Probabilistic forecasting with autoregressive recurrent networks.
\newblock \emph{International journal of forecasting}, 36\penalty0 (3):\penalty0 1181--1191, 2020.

\bibitem[Santos \& Papa(2022)Santos and Papa]{santos2022avoiding}
Claudio Filipi Gon{\c{c}}alves~Dos Santos and Jo{\~a}o~Paulo Papa.
\newblock Avoiding overfitting: A survey on regularization methods for convolutional neural networks.
\newblock \emph{ACM Computing Surveys (CSUR)}, 54\penalty0 (10s):\penalty0 1--25, 2022.

\bibitem[Satorras et~al.(2022)Satorras, Rangapuram, and Januschowski]{satorras2022multivariate}
Victor~Garcia Satorras, Syama~Sundar Rangapuram, and Tim Januschowski.
\newblock Multivariate time series forecasting with latent graph inference.
\newblock \emph{arXiv preprint arXiv:2203.03423}, 2022.

\bibitem[Seo et~al.(2018)Seo, Defferrard, Vandergheynst, and Bresson]{seo2018structured}
Youngjoo Seo, Micha{\"e}l Defferrard, Pierre Vandergheynst, and Xavier Bresson.
\newblock {Structured sequence modeling with graph convolutional recurrent networks}.
\newblock In \emph{International Conference on Neural Information Processing}, pp.\  362--373. Springer, 2018.

\bibitem[Shang et~al.(2021)Shang, Chen, and Bi]{shang2021discrete}
Chao Shang, Jie Chen, and Jinbo Bi.
\newblock Discrete graph structure learning for forecasting multiple time series.
\newblock \emph{International Conference on Learning Representations}, 2021.

\bibitem[Shao et~al.(2022)Shao, Zhang, Wang, Wei, and Xu]{shao2022spatial}
Zezhi Shao, Zhao Zhang, Fei Wang, Wei Wei, and Yongjun Xu.
\newblock Spatial-temporal identity: A simple yet effective baseline for multivariate time series forecasting.
\newblock In \emph{Proceedings of the 31st ACM International Conference on Information \& Knowledge Management}, pp.\  4454--4458, 2022.

\bibitem[Shih et~al.(2019)Shih, Sun, and Lee]{shih2019temporal}
Shun-Yao Shih, Fan-Keng Sun, and Hung-yi Lee.
\newblock Temporal pattern attention for multivariate time series forecasting.
\newblock \emph{Machine Learning}, 108:\penalty0 1421--1441, 2019.

\bibitem[Smyl(2020)]{smyl2020hybrid}
Slawek Smyl.
\newblock {A hybrid method of exponential smoothing and recurrent neural networks for time series forecasting}.
\newblock \emph{International Journal of Forecasting}, 36\penalty0 (1):\penalty0 75--85, 2020.

\bibitem[Srivastava et~al.(2014)Srivastava, Hinton, Krizhevsky, Sutskever, and Salakhutdinov]{srivastava2014dropout}
Nitish Srivastava, Geoffrey Hinton, Alex Krizhevsky, Ilya Sutskever, and Ruslan Salakhutdinov.
\newblock Dropout: a simple way to prevent neural networks from overfitting.
\newblock \emph{The journal of machine learning research}, 15\penalty0 (1):\penalty0 1929--1958, 2014.

\bibitem[Takada \& Fujisawa(2020)Takada and Fujisawa]{takada2020transfer}
Masaaki Takada and Hironori Fujisawa.
\newblock Transfer learning via $\ell_1$ regularization.
\newblock \emph{Advances in Neural Information Processing Systems}, 33:\penalty0 14266--14277, 2020.

\bibitem[Tian \& Zhang(2022)Tian and Zhang]{tian2022comprehensive}
Yingjie Tian and Yuqi Zhang.
\newblock A comprehensive survey on regularization strategies in machine learning.
\newblock \emph{Information Fusion}, 80:\penalty0 146--166, 2022.

\bibitem[Tibshirani(1996)]{tibshirani1996regression}
Robert Tibshirani.
\newblock Regression shrinkage and selection via the lasso.
\newblock \emph{Journal of the Royal Statistical Society Series B: Statistical Methodology}, 58\penalty0 (1):\penalty0 267--288, 1996.

\bibitem[Van~Rossum \& Drake(2009)Van~Rossum and Drake]{python}
Guido Van~Rossum and Fred~L. Drake.
\newblock \emph{Python 3 Reference Manual}.
\newblock CreateSpace, Scotts Valley, CA, 2009.
\newblock ISBN 1441412697.

\bibitem[Vaswani et~al.(2017)Vaswani, Shazeer, Parmar, Uszkoreit, Jones, Gomez, Kaiser, and Polosukhin]{vaswani2017attention}
Ashish Vaswani, Noam Shazeer, Niki Parmar, Jakob Uszkoreit, Llion Jones, Aidan~N Gomez, {\L}ukasz Kaiser, and Illia Polosukhin.
\newblock Attention is all you need.
\newblock \emph{Advances in neural information processing systems}, 30, 2017.

\bibitem[Wang \& Niepert(2019)Wang and Niepert]{wang2019state}
Cheng Wang and Mathias Niepert.
\newblock State-regularized recurrent neural networks.
\newblock In \emph{International Conference on Machine Learning}, pp.\  6596--6606. PMLR, 2019.

\bibitem[Wang et~al.(2018)Wang, Xia, Zhao, Bian, Qin, Liu, and Liu]{wang2018dual}
Yijun Wang, Yingce Xia, Li~Zhao, Jiang Bian, Tao Qin, Guiquan Liu, and Tie-Yan Liu.
\newblock Dual transfer learning for neural machine translation with marginal distribution regularization.
\newblock In \emph{Proceedings of the AAAI Conference on Artificial Intelligence}, volume~32, 2018.

\bibitem[Wang et~al.(2019)Wang, Smola, Maddix, Gasthaus, Foster, and Januschowski]{wang2019deep}
Yuyang Wang, Alex Smola, Danielle Maddix, Jan Gasthaus, Dean Foster, and Tim Januschowski.
\newblock {Deep factors for forecasting}.
\newblock In \emph{International conference on machine learning}, pp.\  6607--6617. PMLR, 2019.

\bibitem[Wu et~al.(2021)Wu, Zhuang, Labbe, and Sun]{wu2021inductive}
Yuankai Wu, Dingyi Zhuang, Aurelie Labbe, and Lijun Sun.
\newblock Inductive graph neural networks for spatiotemporal kriging.
\newblock In \emph{Proceedings of the AAAI Conference on Artificial Intelligence}, volume~35, pp.\  4478--4485, 2021.

\bibitem[Wu et~al.(2019)Wu, Pan, Long, Jiang, and Zhang]{wu2019graph}
Zonghan Wu, Shirui Pan, Guodong Long, Jing Jiang, and Chengqi Zhang.
\newblock Graph wavenet for deep spatial-temporal graph modeling.
\newblock \emph{Proceedings of the Twenty-Eighth International Joint Conference on Artificial Intelligence}, 2019.

\bibitem[Xiao et~al.(2024)Xiao, Liu, and Dyer]{xiao2023gaformer}
Jingyun Xiao, Ran Liu, and Eva~L Dyer.
\newblock Gaformer: Enhancing timeseries transformers through group-aware embeddings.
\newblock In \emph{International Conference on Learning Representations}, 2024.

\bibitem[Yadan(2019)]{Yadan2019Hydra}
Omry Yadan.
\newblock Hydra - a framework for elegantly configuring complex applications.
\newblock Github, 2019.
\newblock URL \url{https://github.com/facebookresearch/hydra}.

\bibitem[Yang et~al.(2019)Yang, Dai, Yang, Carbonell, Salakhutdinov, and Le]{yang2019xlnet}
Zhilin Yang, Zihang Dai, Yiming Yang, Jaime Carbonell, Russ~R Salakhutdinov, and Quoc~V Le.
\newblock {XLNet: Generalized autoregressive pretraining for language understanding}.
\newblock \emph{Advances in neural information processing systems}, 32, 2019.

\bibitem[Yin et~al.(2022)Yin, Li, Shen, Qi, and Yin]{yin2022nodetrans}
Xueyan Yin, Feifan Li, Yanming Shen, Heng Qi, and Baocai Yin.
\newblock Nodetrans: A graph transfer learning approach for traffic prediction.
\newblock \emph{arXiv preprint arXiv:2207.01301}, 2022.

\bibitem[Ying(2019)]{ying2019overview}
Xue Ying.
\newblock An overview of overfitting and its solutions.
\newblock In \emph{Journal of physics: Conference series}, volume 1168, pp.\  022022. IOP Publishing, 2019.

\bibitem[Yu et~al.(2018)Yu, Yin, and Zhu]{yu2018spatio}
Bing Yu, Haoteng Yin, and Zhanxing Zhu.
\newblock Spatio-temporal graph convolutional networks: a deep learning framework for traffic forecasting.
\newblock In \emph{Proceedings of the 27th International Joint Conference on Artificial Intelligence}, 2018.

\bibitem[Zaheer et~al.(2017)Zaheer, Kottur, Ravanbakhsh, Poczos, Salakhutdinov, and Smola]{zaheer2017deep}
Manzil Zaheer, Satwik Kottur, Siamak Ravanbakhsh, Barnabas Poczos, Russ~R Salakhutdinov, and Alexander~J Smola.
\newblock {Deep sets}.
\newblock \emph{Advances in Neural Information Processing Systems}, 30, 2017.

\bibitem[Zaremba et~al.(2014)Zaremba, Sutskever, and Vinyals]{zaremba2014recurrent}
Wojciech Zaremba, Ilya Sutskever, and Oriol Vinyals.
\newblock Recurrent neural network regularization.
\newblock \emph{arXiv preprint arXiv:1409.2329}, 2014.

\bibitem[Zhang et~al.(2022)Zhang, Zeman, Tsiligkaridis, and Zitnik]{zhang2022graph}
Xiang Zhang, Marko Zeman, Theodoros Tsiligkaridis, and Marinka Zitnik.
\newblock {Graph-Guided Network For Irregularly Sampled Multivariate Time Series}.
\newblock In \emph{International Conference on Learning Representations, ICLR}, 2022.

\bibitem[Zheng et~al.(2015)Zheng, Yi, Li, Li, Shan, Chang, and Li]{zheng2015forecasting}
Yu~Zheng, Xiuwen Yi, Ming Li, Ruiyuan Li, Zhangqing Shan, Eric Chang, and Tianrui Li.
\newblock Forecasting fine-grained air quality based on big data.
\newblock In \emph{Proceedings of the 21th ACM SIGKDD international conference on knowledge discovery and data mining}, pp.\  2267--2276, 2015.

\bibitem[Zhou et~al.(2021)Zhou, Vani, Larochelle, and Courville]{zhou2021fortuitous}
Hattie Zhou, Ankit Vani, Hugo Larochelle, and Aaron Courville.
\newblock Fortuitous forgetting in connectionist networks.
\newblock In \emph{International Conference on Learning Representations}, 2021.

\bibitem[Zippenfenig(2023)]{openmeteo}
Patrick Zippenfenig.
\newblock Open-meteo.com weather api.
\newblock \url{https://open-meteo.com/}, 2023.

\end{thebibliography}
\bibliographystyle{tmlr}

\newpage
\appendix
\section*{Appendix}

\section{Datasets}\label{app:datasets}
\begin{table}[!ht]
            \begin{center}
                \begin{small}
                    \begin{sc}
                        \resizebox{\textwidth}{!}{%
    \begin{tabular}{l c c c c c c || c c}
    \toprule
    DATASETS & \# Time series & Time steps & Channels & Connectivity & Edges & Sampling Rate & Time window & Horizon \\
    \midrule
    \gls{la}  & 207  & 34,272 & 1  & Directed   & 1515 & 5 minutes  & 12 & 12 \\
    \gls{bay} & 325  & 52,128 & 1  & Directed   & 2369 & 5 minutes  & 12 & 12 \\
    \gls{cer}    & 485  & 25,728 & 1  & Directed   & 4365 & 30 minutes & 48 & 6 \\
    \gls{air}      & 437  & 8,760  & 1  & Undirected & 2699 & 1 hour     & 24 & 3 \\
    \gls{climate}    & 235  & 10,958 & 10 & NA        & 2699 & 1 day      & 14 & 3 \\
    \gls{engrad}   & 487  & 26,304 & 5  & NA        & 2699 & 1 hour     & 24 & 6 \\
    \gls{pems3}   & 358  & 26,208 & 1  & Directed   & 546  & 5 minutes  & 12 & 12 \\
    \gls{pems4}   & 307  & 16,992 & 1  & Directed   & 340  & 5 minutes  & 12 & 12 \\
    \gls{pems7}   & 883  & 28,224 & 1  & Directed   & 866  & 5 minutes  & 12 & 12 \\
    \gls{pems8}   & 170  & 17,856 & 1  & Directed   & 277  & 5 minutes  & 12 & 12 \\
    \bottomrule
    \end{tabular}
    \caption{Dataset details.}
    \label{tab:datasets}
                }
                \end{sc}
            \end{small}
        \end{center}    
\end{table}

\paragraph{\gls{la}} traffic data from~\citet{li2018diffusion}. Traffic readings are from different loop detectors on highways in the Los Angeles County. Licensed under Attribution 4.0 International (CC BY 4.0).

\paragraph{\gls{bay}} traffic data from~\citet{li2018diffusion}. Traffic readings are from different loop detectors on highways in the Los Angeles County. Licensed under Attribution 4.0 International (CC BY 4.0).

\paragraph{\gls{cer}} Electric load data from~\citep{ceren}. Data encompass energy consumption readings from smart meters in small and medium enterprises, collected in the context of the Commission for Energy Regulation (CER) Smart Metering Project. Data access can be requested through \url{https://www.ucd.ie/issda/data/commissionforenergyregulationcer/}

\paragraph{\gls{air}} Air quality data from~\citet{zheng2015forecasting}. The dataset collects measurements of the \textit{PM2.5} pollutant from air quality stations in 43 Chinese cities and is available at \url{https://www.microsoft.com/en-us/research/publication/forecasting-fine-grained-air-quality-based-on-big-data/}

\paragraph{\gls{climate}} satellite daily climatic dataset from~\citep{de2024graph}. Data were obtained from the POWER Project’s Daily 2.3.5 version on 2023/02/26 and sampled in the correspondence of the 235 world capitals. Further information, together with data and API, is available at the project website~(\url{https://power.larc.nasa.gov/}). For daily data, we select the following 10 variables: \textit{mean temperature} ($^{\circ}C$), \textit{temperature range} ($^{\circ}C$), \textit{maximum temperature} ($^{\circ}C)$, \textit{wind speed} ($m/s$), \textit{relative humidity} ($\%$), \textit{precipitation} ($mm$/day), \textit{dew/frost point} ($^{\circ}C$), \textit{cloud amount} ($\%$), \textit{allsky surface shortwave irradiance} ($W/m^2$) \textit{and all-sky surface longwave irradiance} $(W/m^2$). Data extend for 30 years (1991 to 2022).

\paragraph{\gls{engrad}} hourly climatic dataset from~\citep{marisca2024graph}. The measurements are provided by \url{https://open-meteo.com}~\citep{openmeteo} and licensed under Attribution 4.0 International (CC BY 4.0).
Data are collected on a grid in correspondence with cities in England. The variables correspond to \textit{air temperature} at 2 meters above ground ($^{\circ}C$); \textit{relative humidity} at 2 meters above ground ($\%$); summation of total \textit{precipitation} (rain, showers, snow) during the preceding hour ($mm$); total cloud cover (\%); global horizontal irradiation ($W/m^2$). Data extend for 3 years (2018 to 2020). 

\paragraph{\gls{pems3}, \gls{pems4}, \gls{pems7}, and \gls{pems8}} datasets from ~\citet{guo2021learning} collect traffic detector data from 4 districts in California provided by Caltrans Performance Measurement System (PeMS). Data are aggregated into 5-minutes intervals.

All datasets were split \textit{70\%/10\%/20\%} into \textit{train}, \textit{validation} and \textit{test} along the temporal axis.
Datasets with a high number of missing values, i.e., \gls{la}, \gls{bay} and \gls{air}, have a binary mask concatenated to the input, indicating if the corresponding value has been artificially imputed.  
Note that, unless differently specified, used datasets are public domain.
Where possible, we pointed to the original source of the data or to a meaningful reference.

\section{Models}\label{app:models}
In this section we describe the models used in our study.
Note that all models used a fixed hidden size $d_h$ for all layers.
For all the architectures, the \textsc{Encoder}~\ref{eq:encoder} is parametrized by a linear layer, while the \textsc{Decoder} is a 1-layer \gls{mlp} followed by $H$ parallel linear layers, each decoding a different step in the forecasting horizon.
The \gls{rnn} model is implemented by means of a 1-layer \gls{gru}~\citep{cho2014learning} shared among all sequences.
On top of the same \gls{gru} architecture, for the \gls{stgnn} model, we employ 2 layers of message passing defined as
\begin{align}
    \mathbf{m}^{j\rightarrow i} &= \mathbf{W}_2\xi \left(\mathbf{W}_1\left[\vh^i, \vh^j, a_{ji}\right]\right), \qquad \alpha^{j \rightarrow i} = \eta\left(\mathbf{W}_0 \mathbf{m}^{j \rightarrow i}\right),\\
    \mathbf{\tilde{h}}^i &= \xi \left(\mathbf{W}_3 \vh^i + \sum_{j \in \mathcal{N}\left(i\right)}\{\alpha^{j \rightarrow i} \odot \mathbf{m}^{j \rightarrow i}\}\right),
\end{align}
where $\mathbf{W}_0 \in \sR^{d_h \times d_h}$, $\mathbf{W}_1 \in \sR^{d_h \times \left(2d_h + 1\right)}$, $\mathbf{W}_2 \in \sR^{d_h \times d_h}$ and $\mathbf{W}_3 \in \sR^{d_h \times d_h}$ are learnable parameters, $\left[\cdot,\cdot\right]$ is the concatenation operator along the feature dimension, $\odot$ is the Hadamard product, $\xi$ denotes the \textit{elu}~\citep{clevert2016elu} activation function and $\eta$ the sigmoid activation function.
Furthermore, $\mathcal{N}\left(i\right)$ denotes the neighbours of the $i$-th sequence, induced by the adjacency matrix $\mathbf{A}$, $a_{ji}$ denotes the weight associated to edge $j \rightarrow i$ and $\vh^i$ and $\vh^j$ denote the hidden features associated with the $i$-th and $j$-th sequences respectively.
Regarding the \gls{statt} model, we use 2 layers of multi-head attention~\citep{vaswani2017attention} taking as input tokens the hidden representations extracted by a temporal encoder which shares the same architecture as the aforementioned \gls{rnn}.
Finally, for consistency with the original experiment~\citep{cini2023taming}, we employ an \gls{stgnn} consisting of a 1-layer \gls{gru} and 2 message passing layer implementing
\begin{equation}
    \mathbf{\tilde{h}}^i = \xi \left(\mathbf{W}_4 \vh^i + \frac{1}{\|\mathcal{N}\left(i\right)\|} \sum_{j \in \mathcal{N}\left(i\right)}\{\mathbf{W}_5 \vh^j\}\right),
    \label{eq:stgnn_imp}
\end{equation}
where $\mathbf{W}_4 \in \sR^{d_h \times d_h}$ and $\mathbf{W}_5 \in \sR^{d_h \times d_h}$ are learnable parameters, while $\|\cdot\|$ is the cardinality operator.

\section{Additional details on forgetting regularization}\label{app:forgetting_details}

For the sake of completeness, in the following, we list some additional details regarding the local forgetting regularization.

\subsection{Embedding-related Encoder/Decoder parameters reset} \label{app:encoder_decoder_reset}

When resetting the local parameters, we also reset the encoder's~(Eq.~\ref{eq:encoder}) and decoder's~(Eq.~\ref{eq:decoder}) parameters (i.e., coefficients of a linear layer) that directly interact with the embeddings' features.
As an illustrative example, consider an $\textsc{Encoder}$ parametrized by a \gls{mlp}, with input linear layer
\begin{equation}
    \vh_t^0 = \left[\mX_{t-1} \Vert \mU_{t-1} \Vert \mE\right] \mW^T + \vb,
\end{equation}
In this case, the encoder's parameters to be reset would correspond to the last $d_e$ columns of the weight matrix $\mW$.
An equivalent behavior is implemented for the decoder's input layer.

\subsection{Forgetting period sensitivity} \label{app:forgetting_sensitivity}

\begin{figure}[t]
    \begin{subfigure}{0.85\textwidth}
        \centering
        \includegraphics[width=\textwidth]{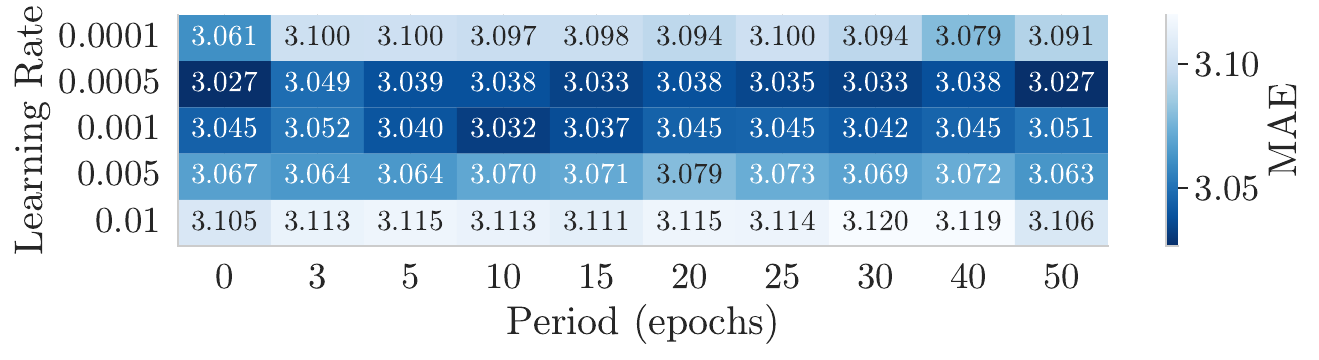}
        \vspace{-.7cm}
        \caption{\gls{la}}
        \label{fig:period_sensitivity_metrla}
    \end{subfigure}
    \begin{subfigure}{0.85\textwidth}
        \centering
        \includegraphics[width=\textwidth]{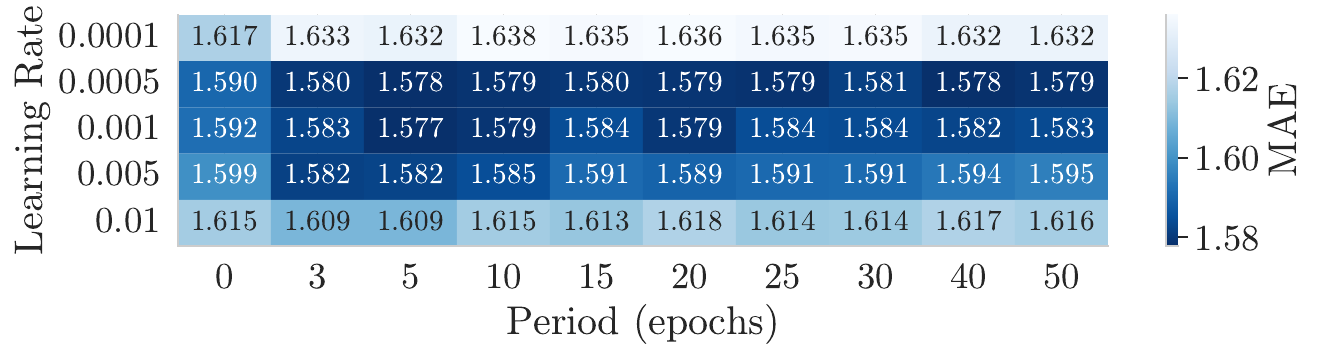}
        \vspace{-.7cm}
        \caption{\gls{bay}}
        \label{fig:period_sensitivity_pemsbay}
    \end{subfigure}
    \begin{subfigure}{0.85\textwidth}
        \centering
        \includegraphics[width=\textwidth]{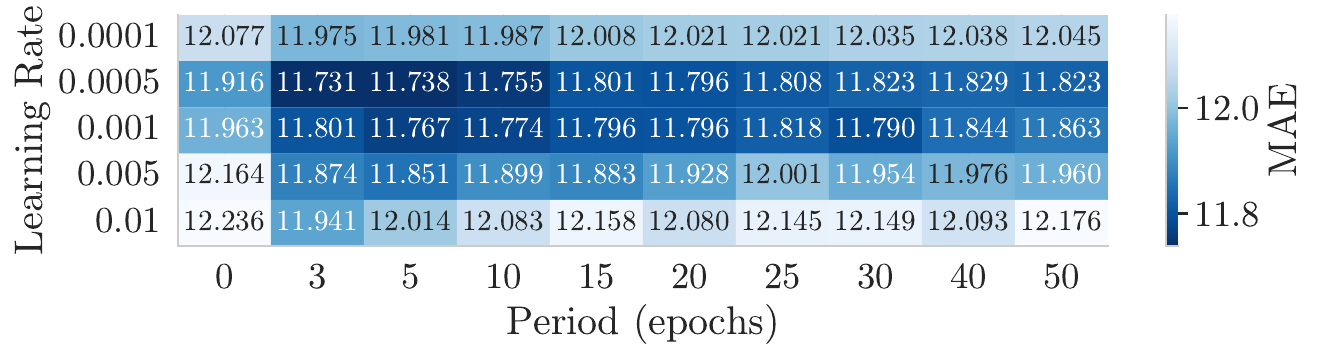}
        \vspace{-.7cm}
        \caption{\gls{air}}
        \label{fig:period_sensitivity_airquality}
    \end{subfigure}
    \caption{Sensitivity to forgetting period $k$ across different learning rates and datasets (5 runs, \gls{stgnn}). A period of $0$ indicates no regularization.}
    \label{fig:period_sensitivity}
\end{figure}
The introduced \textit{forgetting} regularization has two hyperparameters: the reset period $k$ and the halting epoch.
While the latter can be determined automatically, by monitoring the validation error before each reset, the former should be set empirically.
To provide insight on the degree to which the selection of $k$ can impact regularization performance, we perform a sensitivity study on $3$ datasets: \gls{la}, \gls{bay} and \gls{air}.
The results are shown in Fig.~\ref{fig:period_sensitivity}, where we report the \acrfull{mae} with varying learning rates and values of $k$.
We can see that performance is relatively stable in regard to changes in $k$, suggesting that hyperparameter search can be limited to few values (e.g., $3$ for a short period and $30$ for a long one).
Another possibility, to avoid searches, is to select the value by hand, after inspecting the learning curves of the unregularized model, in order to select a reset period that allows performance recovery.

\subsection{Forgetting warm-up}

To avoid instabilities at the beginning of training, particularly when using short reset periods, the forgetting routine can be initiated after a warm-up period, which can be set empirically to a few epochs, as commonly done with learning rate warm-up procedures or with other regularizations~\citep{cini2023taming}. 

\section{Detailed experimental setting}\label{app:details}

\paragraph{Reproducibility}
Python code to reproduce the experiments is available online\footnote{\url{https://github.com/LucaButera/TS-embedding-regularization}}. 
Datasets will be automatically downloaded or access can be requested to the authors, aside from \gls{cer} and \gls{engrad}. Such datasets can be obtained by contacting the original papers' authors (\gls{engrad}~\citep{marisca2024graph}) or by requesting data access to the appropriate authority (\gls{cer}~\citep{ceren}). To specifically reproduce the experiments in Tab~\ref{tab:transfer}, regarding the columns for \textit{clustering} and \textit{variational} regularizations, access to the code can be requested to the original authors of~\citet{cini2023taming}. Appendix~\ref{app:datasets} contains pointers to request such access. Note that randomized operations in the code are controlled by fixed seeds for reproducibility purposes. 

\paragraph{Shared settings}
All the models were trained with the Adam optimizer~\citep{kingma2015adam}, with a batch size of 64  and up to 300 batches per epoch.
We used the \textit{Python}~\citep{python} programming language, leveraging \textit{Torch Spatiotemporal}~\citep{tsl}, \textit{Pytorch}~\citep{paszke2019pytorch} and \textit{Pytorch Lightning}~\citep{Falcon_PyTorch_Lightning_2019} to implement all the experiments.
Experiments were scheduled and logged by leveraging \textit{Hydra}~\citep{Yadan2019Hydra} and \textit{Weights and Biases}~\citep{wandb}. 
The \acrfull{mae} was used as loss function in all experiments.
For \textit{variational} and \textit{clustering} regularizations additional hyperparameters, aside from the regularization strength, i.e., $\lambda_{var}$, $\lambda_{clst}$, were taken from the original paper. 
Unless specified, we employed and embedding size $d_e = 32$ for all models.
Moreover, all datasets used temporal encodings as additional covariates, in particular, a one-hot encoding of the weekday and sinus and cosinus encodings with daily period.

\paragraph{Experiment specific hyperparameters}
For the experiments in Tab.~\ref{tab:pred_performance}, optimal hyperparameters, i.e., learning rate $lr$ and hidden size $d_{h}$, for each model and dataset, were found via a grid-search over $lr \in [0.00025, 0.00075, 0.0015, 0.003]$ and $d_h \in [32, 64, 128, 256]$, with constraints caused by GPU memory capacity in some settings. Before the grid search for regularized models, regularization's hyperparameters $\lambda_{l1}$, $\lambda_{l2}$, $\lambda_{var}$ and $\lambda_{clst}$, were set according to the following procedure. For each model and dataset (we here consider three datasets: \gls{la}, \gls{bay} and \gls{air}), the validation error is computed for regularization strength values in $\left[0.01, 0.001, 0.0001, 0.00001\right]$, while keeping the $lr$ and $d_{h}$ fixed to the optimal values for the respective un-regularized model. For each regularization hyperparameter, the corresponding validation results are ranked and the average rank is computed across models and datasets. The parameter with the top rank is selected and kept fixed. This resulted in: weight of L2 $\lambda_{l2} = 0.0001$, weight of L1 $\lambda_{l1} = 0.00001$, weight of \textit{variational} regularization $\lambda_{var} = 0.00005$ and weight of \textit{clustering} regularization $\lambda_{clst} = 0.0005$. 
Dropout's probability of dropping a connection was set to $p=0.5$, which is a middle ground between soft and aggressive dropout. 
Moreover, we set the resetting period of \textit{forgetting} to $k=20$ epochs with $30$ epochs of warm-up. The analysis in Sec.~\ref{app:forgetting_sensitivity} shows how the method is not particularly sensitive in this regard. Forgetting was halted after $150$ epochs. 
Each model was trained for $200$ epochs and the best validation parameters were used for testing. 

For the experiments in Tab.~\ref{tab:transfer}, we used the \gls{stgnn} in Eq.~\ref{eq:stgnn_imp} and set $d_{h} = 64$, $\lambda_{var} = 0.05$ and $\lambda_{clst} = 0.5$ as in~\citet{cini2023taming}, while for the other regularizations we used the same hyperparameters as in Tab.~\ref{tab:pred_performance}. Furthermore, we used a learning rate $lr = 0.005$ during training and $lr = 0.001$ during fine-tuning. The training lasted up to $150$ epochs with $50$ epochs of early stopping patience, while fine-tuning lasted up to $1000$ epochs with $100$ epochs of patience. No temporal encodings were used in this setting. 

For experiments in Sec.~\ref{subsec:overfit} and Sec.~\ref{subsec:perturbation} we used set $d_h = 64$ and $lr=0.00075$. Regularization's hyperparameters were set to the values found for Tab.~\ref{tab:pred_performance}.
Training lasted up to $300$ epochs, early stopping patience was set to $100$, while forgetting was halted after $100$ epochs. 
Note that the \textit{forgetting} regularization ignores early stopping for the whole duration before being halted; this is to avoid triggering early stopping as a byproduct of the parameter resetting itself. 

\paragraph{Metrics} For univariate datasets, we consider the \acrfull{mae}, as this is a standard choice across the related literature. For multivariate datasets, i.e., \gls{climate} and \gls{engrad}, following \citet{de2024graph}, we compute the \acrfull{mre} independently for each channel, then, take the channel-wise average to obtain a unique final score, which we termed \acrfull{mmre} in Tab.~\ref{tab:pred_performance}.

\paragraph{Computing resources}
Experiments were run on A100 and A5000 NVIDIA GPUs. 
The vast majority of the experiments conducted in our work can be easily run on moderate GPU hardware, with at least $8$ GBs of VRAM.
However, not that some experiments, may require more.
For instance, the transfer experiments require at least $20$ GBs of VRAM on some of the benchmark dataset.
Moreover, high hidden size configurations of \gls{stgnn} and \gls{statt} models, required up to $40$ GBs of VRAM.
In general, a single run (i.e., training one model on one dataset), given hardware that fulfills the memory requirements, takes from 30 minutes to 3 hours, depending on the specific configuration.

\subsection{Embeddings perturbation} \label{app:perturbation}
Here we provide a more formal definition of the perturbations adopted in Sec.~\ref{subsec:perturbation}. In particular, considering $N$ embedding vectors:
\begin{itemize}
\item \textit{Noise} $\sigma$ consists in adding zero-mean gaussian noise, from an isotropic multivariate normal distribution, to the embeddings. Formally, this refers to substituting each embedding $\mathbf{e}^i$  with $\mathbf{e}^i + \epsilon$, where $\epsilon \sim \mathcal{N}\left(\mathbf{0}, diag\left(\sigma^2\right)\right)$.
\item \textit{Rearrange} refers to randomly reassigning the embedding vectors to different time-series in the collection. Formally, this means substituting each embedding $\mathbf{e}^i$ with an embedding $\mathbf{e}^j$, where $j \sim Multinomial\left(\{0, ..., N - 1\}\right)$ is a sample from a multinomial distribution over the embedding indices, sampled without repetition.
\item \textit{Mean} refers to setting each embedding to the mean values across embeddings themselves. Formally, replacing each embedding $\mathbf{e}^i$ with $\frac{1}{N}\sum_{j = 0}^N\mathbf{e}^j$.
\item \textit{Sampled} consists in estimating the sample gaussian distribution of the embeddings, and then replacing each embedding with a draw from such distribution. Formally, this entails estimating the gaussian's parameters as $\mu_e = \frac{1}{N}\sum_{i = 0}^{N - 1}\mathbf{e}^i$ and $\sigma^2_e = \frac{1}{N - 1}\sum_{i=0}^{N - 1}\left(\mathbf{e}^i - \mu\right)$. Then we sample values for each embedding as $\mathbf{e}^i \sim \mathcal{N}\left(\mu_e, \sigma^2_e\right)$.
\end{itemize}
Note that, to avoid penalizing regularizations that are sensible to the weights' magnitude ~(i.e., L1, L2), we consider perturbations that do not change the scale of the learned representations.
As this does not hold true for the \textit{Noise} $\sigma$ perturbation, since it can potentially affect the embeddings magnitude, depending on the standard deviation, we selected most values to be within reason relative to the embeddings magnitude itself. Nonetheless, we also employed a few more extreme values, in order to observe their effects.

\section{Effect on simple global models}\label{app:mlp}

The purpose of model regularization is, usually, to limit model capacity by means of additional constraints and, in turn, obtain models that generalize better.
This implies that excessive regularization might result in a degradation of performance.
In principle, regularization of local parameters should not have a negative impact on underparametrized global models, as their parameters are not constrained directly.
To evaluate this, we employ a simple 2-layer \gls{mlp} as our global model~(Eq.~\ref{eq:stp}), and train it with the same settings as in Sec.~\ref{subsec:benchmarks}.

\begin{table}[t]
\caption{Forecasting error in transductive setting for \acrshort{mlp} model (hidden size 64-128) (5 runs, $\pm 1std$). Equal to/better than unregularized (\textsc{+ Emb}) in bold. Best in red. `\textsc{+ Regularization}' denotes the addition of that specific regularization only, on top of `\textsc{+ Emb.}'.}
\label{tab:pred_perf_mlp}
\begin{center}
\begin{small}
\setlength{\tabcolsep}{2pt}
\setlength{\aboverulesep}{0.25pt}
\setlength{\belowrulesep}{0.25pt}
\begin{sc}
\vspace{-.3cm}\resizebox{\textwidth}{!}{%
\begin{tabular}{l| c | c | c | c | c | c}
\toprule[1pt]
\multicolumn{1}{c|}{\sc Dataset} & \multicolumn{1}{c|}{\gls{la}} & \multicolumn{1}{c|}{\gls{bay}} & \multicolumn{1}{c|}{\gls{cer}} & \multicolumn{1}{c|}{\gls{air}} & \multicolumn{1}{c|}{\gls{climate}} & \multicolumn{1}{c}{\gls{engrad}}\\
\cmidrule(l{15pt}r{15pt}){1-1} \cmidrule{2-7}
\multicolumn{1}{c|}{\sc Model} & {\acrshort{mae} $\downarrow$} & {\acrshort{mae} $\downarrow$} & {\acrshort{mae} $\downarrow$} & {\acrshort{mae} $\downarrow$} & {\acrshort{mmre} $\downarrow$} & {\acrshort{mmre} $\downarrow$}\\
\arrayrulecolor{black}\midrule[1pt]
\acrshort{mlp} (64) & $3.580_{\pm 0.005}$          & $1.808_{\pm 0.002}$          & $0.4658_{\pm 0.0008}$ & $\mathbf{13.398}_{\pm 0.044}$ & $19.86_{\pm 0.01}$ & $\mathbf{31.38}_{\pm 0.13}$\\
+ Emb.   & $3.153_{\pm 0.012}$          & $1.619_{\pm 0.005}$          & $0.4233_{\pm 0.0007}$ & $13.521_{\pm 0.082}$ & $19.55_{\pm 0.01}$ & $31.40_{\pm 0.14}$\\
\arrayrulecolor{black!30}\midrule
+ L1     & $\mathbf{3.150}_{\pm 0.005}$ & $\mathbf{1.614}_{\pm 0.006}$ & $0.4236_{\pm 0.0008}$ & $\mathbf{13.392}_{\pm 0.067}$ & $19.56_{\pm 0.01}$ & $\mathbf{31.32}_{\pm 0.04}$\\
+ L2     & $\mathbf{3.147}_{\pm 0.007}$ & $\mathbf{1.613}_{\pm 0.004}$ & $0.4242_{\pm 0.0010}$ & $\mathbf{13.389}_{\pm 0.042}$ & $19.57_{\pm 0.02}$ & $\mathbf{31.21}_{\pm 0.09}$\\
+ Clust. & $\mathbf{3.152}_{\pm 0.014}$ & $1.625_{\pm 0.009}$          & $0.4236_{\pm 0.0012}$ & $\mathbf{13.478}_{\pm 0.070}$ & ${\color{BrownRed}\mathbf{19.54}}_{\pm 0.01}$ & $\mathbf{31.25}_{\pm 0.14}$\\
+ Drop.  & $3.193_{\pm 0.003}$          & $1.630_{\pm 0.006}$          & $0.4317_{\pm 0.0009}$ & ${\color{BrownRed}\mathbf{13.328}}_{\pm 0.036}$ & $19.62_{\pm 0.01}$ & ${\color{BrownRed}\mathbf{31.20}}_{\pm 0.07}$\\
+ Vari.  & ${\color{BrownRed}\mathbf{3.139}}_{\pm 0.011}$ & ${\color{BrownRed}\mathbf{1.610}}_{\pm 0.001}$ & ${\color{BrownRed}\mathbf{0.4216}}_{\pm 0.0011}$ & $\mathbf{13.337}_{\pm 0.020}$ & ${\color{BrownRed}\mathbf{19.54}}_{\pm 0.01}$ & $\mathbf{31.32}_{\pm 0.03}$\\
+ Forg.  & $3.157_{\pm 0.013}$          & $\mathbf{1.617}_{\pm 0.007}$ & $0.4271_{\pm 0.0012}$ & $\mathbf{13.472}_{\pm 0.099}$ & $19.56_{\pm 0.02}$ & $\mathbf{31.34}_{\pm 0.09}$\\
\arrayrulecolor{black}\midrule[1pt]
\acrshort{mlp} (128)& $3.571_{\pm 0.002}$          & $1.798_{\pm 0.003}$          & $0.4557_{\pm 0.0003}$ & $\mathbf{13.390}_{\pm 0.044}$ & $19.78_{\pm 0.01}$ & $31.44_{\pm 0.04}$\\
+ Emb.   & $3.157_{\pm 0.014}$          & $1.615_{\pm 0.006}$          & $0.4141_{\pm 0.0013}$ & $13.534_{\pm 0.052}$ & ${\color{BrownRed}\mathbf{19.47}}_{\pm 0.01}$ & $31.38_{\pm 0.10}$\\
\arrayrulecolor{black!30}\midrule
+ L1     & $\mathbf{3.152}_{\pm 0.004}$ & $\mathbf{1.604}_{\pm 0.006}$ & $\mathbf{0.4128}_{\pm 0.0007}$ & $\mathbf{13.417}_{\pm 0.078}$ & $19.49_{\pm 0.02}$ & $\mathbf{31.38}_{\pm 0.07}$\\
+ L2     & $\mathbf{3.150}_{\pm 0.015}$ & ${\color{BrownRed}\mathbf{1.600}}_{\pm 0.006}$ & $\mathbf{0.4117}_{\pm 0.0009}$ & $\mathbf{13.450}_{\pm 0.059}$ & $19.48_{\pm 0.01}$ & $31.65_{\pm 0.17}$\\
+ Clust. & $\mathbf{3.143}_{\pm 0.017}$ & $\mathbf{1.603}_{\pm 0.003}$ & $\mathbf{0.4134}_{\pm 0.0009}$ & $13.612_{\pm 0.119}$ & ${\color{BrownRed}\mathbf{19.47}}_{\pm 0.01}$ & $31.42_{\pm 0.06}$\\
+ Drop.  & $3.195_{\pm 0.008}$          & $1.624_{\pm 0.003}$          & ${\color{BrownRed}\mathbf{0.4101}}_{\pm 0.0003}$ & ${\color{BrownRed}\mathbf{13.320}}_{\pm 0.044}$ & $19.53_{\pm 0.02}$ & $\mathbf{31.35}_{\pm 0.13}$\\
+ Vari.  & ${\color{BrownRed}\mathbf{3.139}}_{\pm 0.017}$ & $\mathbf{1.603}_{\pm 0.005}$ & $\mathbf{0.4109}_{\pm 0.0012}$ & $\mathbf{13.346}_{\pm 0.038}$ & ${\color{BrownRed}\mathbf{19.47}}_{\pm 0.01}$ & ${\color{BrownRed}\mathbf{31.33}}_{\pm 0.07}$\\
+ Forg.  & $\mathbf{3.150}_{\pm 0.014}$ & $1.616_{\pm 0.010}$          & $0.4184_{\pm 0.0008}$ & $\mathbf{13.427}_{\pm 0.058}$ & $19.53_{\pm 0.02}$ & $\mathbf{31.36}_{\pm 0.09}$\\
\arrayrulecolor{black}\bottomrule[1pt]
\end{tabular}
}
\end{sc}
\end{small}
\end{center}             
\end{table}
Tab.~\ref{tab:pred_perf_mlp} reports the obtained results in case of tuned learning rate and hidden size of $64$ and $128$ units.
Compared to models in Tab.~\ref{tab:pred_performance}, the performance of the \gls{mlp} is worse or comparable at best, however, the regularizations relative effectiveness is similar, considering each dataset, at hidden size $128$.
In the extreme case of the \gls{mlp} with hidden size $64$, we can observe additional scenarios in which most regularizations are ineffective (i.e., \gls{cer}).
Nonetheless, we do not observe catastrophic effects on performance.
Notably, with some datasets, i.e., \gls{air} and \gls{engrad}, adding local parameters to the global \gls{mlp} model hurts performance, while regularization allows an effective exploitation of the embeddings.

\section{Effect of combining different regularizations}\label{app:combo}

To further investigate whether different regularization techniques can be combined for improved performance, we applied L2 and dropout on top of the approaches specifically designed for the regularization of local embeddings, i.e., \textit{clustering}, \textit{variational} and \textit{forgetting} regularizations. This covers the most reasonable combinations, as L2 and \textit{dropout} are commonly used in combination with other regularization approaches. In particular, we consider the same transductive setting as in Sec.~\ref{subsec:benchmarks}, and adopt the optimal hyperparameter configuration (see Appendix~\ref{app:details} for details) previously found for \textit{clustering}, \textit{variational}, and \textit{forgetting}, respectively. 

\begin{table}[t]
\caption{Forecasting test error under optimal model size and learning rate from Tab.~\ref{tab:pred_performance} (\acrshort{stgnn}, 5 runs, $\pm 1std$) when adding additional regularizations. Combinations equal to or better than the corresponding base regularization are in bold. The best-performing method within each dataset and base regularization is in red. Regularization methods are always considered applied on top of the `\textsc{+ Emb.}' model.}
\label{tab:combo_performance}
\begin{center}
\begin{small}
\setlength{\tabcolsep}{3.5pt}
\setlength{\aboverulesep}{0.5pt}
\setlength{\belowrulesep}{0.5pt}
\begin{sc}
\resizebox{0.7\textwidth}{!}{%
\begin{tabular}{l| c | c | c }
\toprule[1pt]
\multicolumn{1}{c|}{\sc Dataset} & \multicolumn{1}{c|}{\gls{la}} & \multicolumn{1}{c|}{\gls{bay}} & \multicolumn{1}{c}{\gls{air}}\\
\cmidrule(l{15pt}r{15pt}){1-1} \cmidrule{2-4}
\multicolumn{1}{c|}{\sc Model} & {\acrshort{mae} $\downarrow$} & {\acrshort{mae} $\downarrow$} & {\acrshort{mae} $\downarrow$}\\               
\arrayrulecolor{black}\midrule[1pt]
+ Clust. & $3.025_{\pm .012}$ & $1.580_{\pm .005}$ & $11.876_{\pm .053}$\\
\arrayrulecolor{black!30}\midrule
+ Clust. + L2 & ${\color{BrownRed}\mathbf{3.019}}_{\pm 0.008}$ & $1.581_{\pm 0.008}$ & $\mathbf{11.767}_{\pm 0.023}$ \\
+ Clust. + Drop. & $3.052_{\pm 0.007}$ & ${\color{BrownRed}\mathbf{1.572}}_{\pm 0.009}$ & ${\color{BrownRed}\mathbf{11.712}}_{\pm 0.037}$ \\
\arrayrulecolor{black}\midrule[1pt]
+ Vari. & $\mathbf{3.013}_{\pm .005}$ & $\mathbf{1.566}_{\pm .003}$ & $11.768_{\pm .026}$\\
\arrayrulecolor{black!30}\midrule
+ Vari. + L2 & $3.037_{\pm 0.009}$ & $1.581_{\pm 0.007}$ & $11.789_{\pm 0.022}$ \\
+ Vari. + Drop. & $3.047_{\pm 0.008}$ & $1.576_{\pm 0.004}$ & ${\color{BrownRed}\mathbf{11.675}}_{\pm 0.023}$ \\
\arrayrulecolor{black}\midrule[1pt]
+ Forg. & $3.050_{\pm .017}$ & $1.578_{\pm .006}$ & $11.793_{\pm .040}$\\
\arrayrulecolor{black!30}\midrule
+ Forg. + L2 & $\mathbf{3.045}_{\pm 0.012}$ & $\mathbf{1.576}_{\pm 0.007}$ & $\mathbf{11.764}_{\pm 0.023}$ \\
+ Forg. + Drop. & $3.087_{\pm 0.005}$ & ${\color{BrownRed}\mathbf{1.570}}_{\pm 0.003}$ & ${\color{BrownRed}\mathbf{11.755}}_{\pm 0.028}$ \\
\arrayrulecolor{black}\bottomrule[1pt]
\end{tabular}
}
\end{sc}
\end{small}
\end{center}             
\end{table}
Tab.~\ref{tab:combo_performance} shows the obtained results. We can see that combining different regularization techniques can be beneficial. Noticeably, combining \textit{variational} regularization and dropout yielded a new overall best score on the \gls{air} dataset. However, improvements appear mostly case-dependent, without emerging clear patterns. 

\section{Additional experiments}\label{app:additional_exp}

For completeness, in this section, we provide experimental results in addition to what has been shown in the main paper.

\begin{figure}[t]
     \centering
    \includegraphics[width=\textwidth]{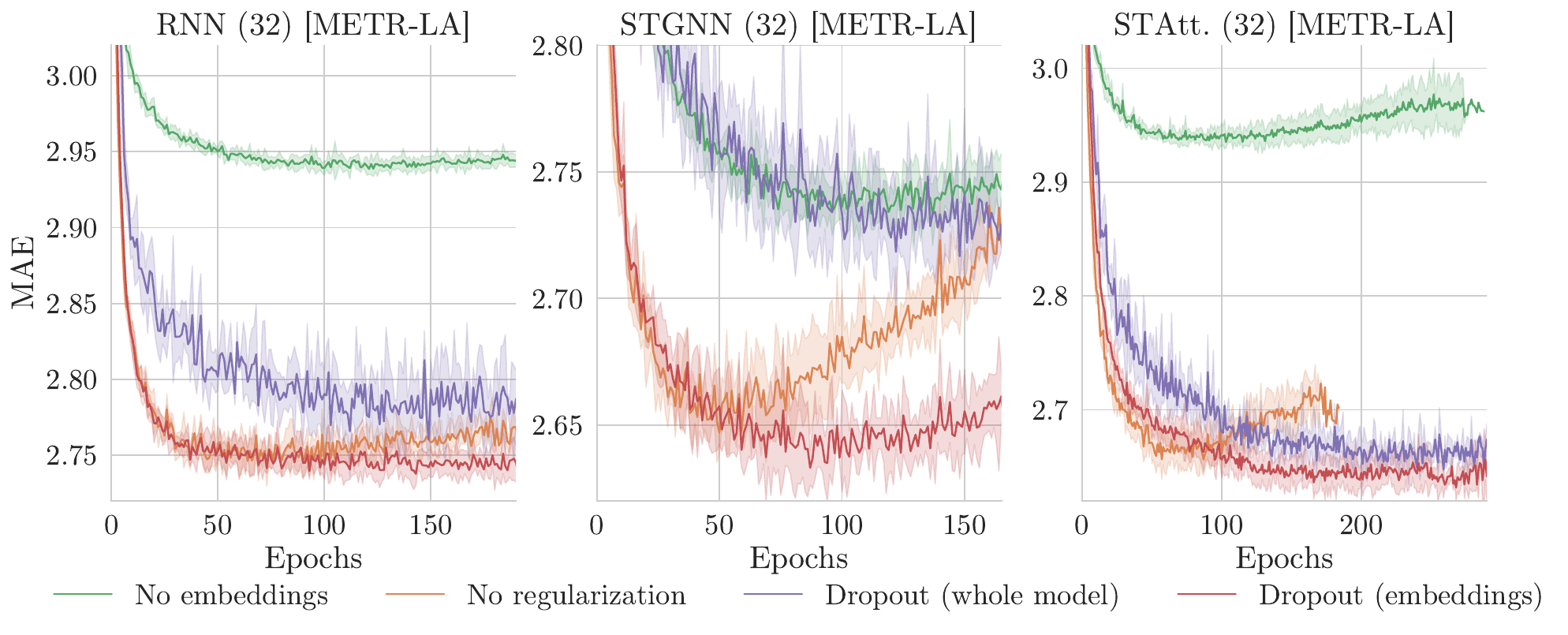}
    \vspace{-.7cm}
        \caption{Validation curves for different model families (5 runs, $\pm 1std$). Curves show the application of dropout regularization at the embedding level or over the entire architecture. Plot names follow the convention \textit{model (embedding size) [dataset]}.}
        \label{fig:app_all_dropout}
\end{figure}
Fig.~\ref{fig:dropout_local_global} shows an example of how regularizing local parameters only can have a different impact than regularizing all parameters on the training of an \gls{stgnn} model. Fig.~\ref{fig:app_all_dropout} complements what has been shown in such an example by providing results for the other two models, i.e., \gls{rnn} and \gls{statt}. A similar pattern emerges for all the considered models, though with different degrees. 

\begin{figure}[t]
    \centering
     \begin{subfigure}{0.95\textwidth}
     \centering
    \includegraphics[width=\textwidth]{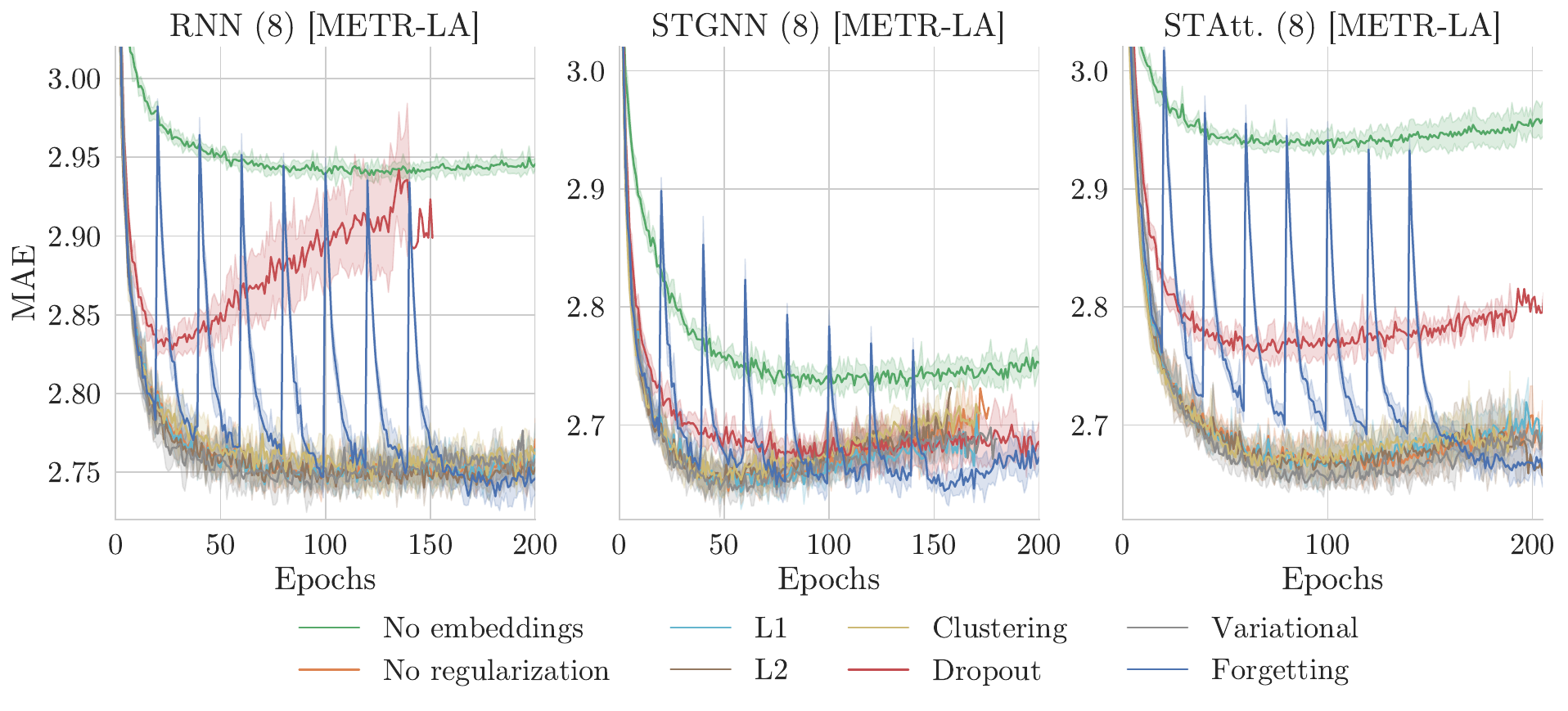}
    \vspace{-.5cm}
    \end{subfigure}
    \begin{subfigure}{0.95\textwidth}
     \centering
    \includegraphics[width=\textwidth]{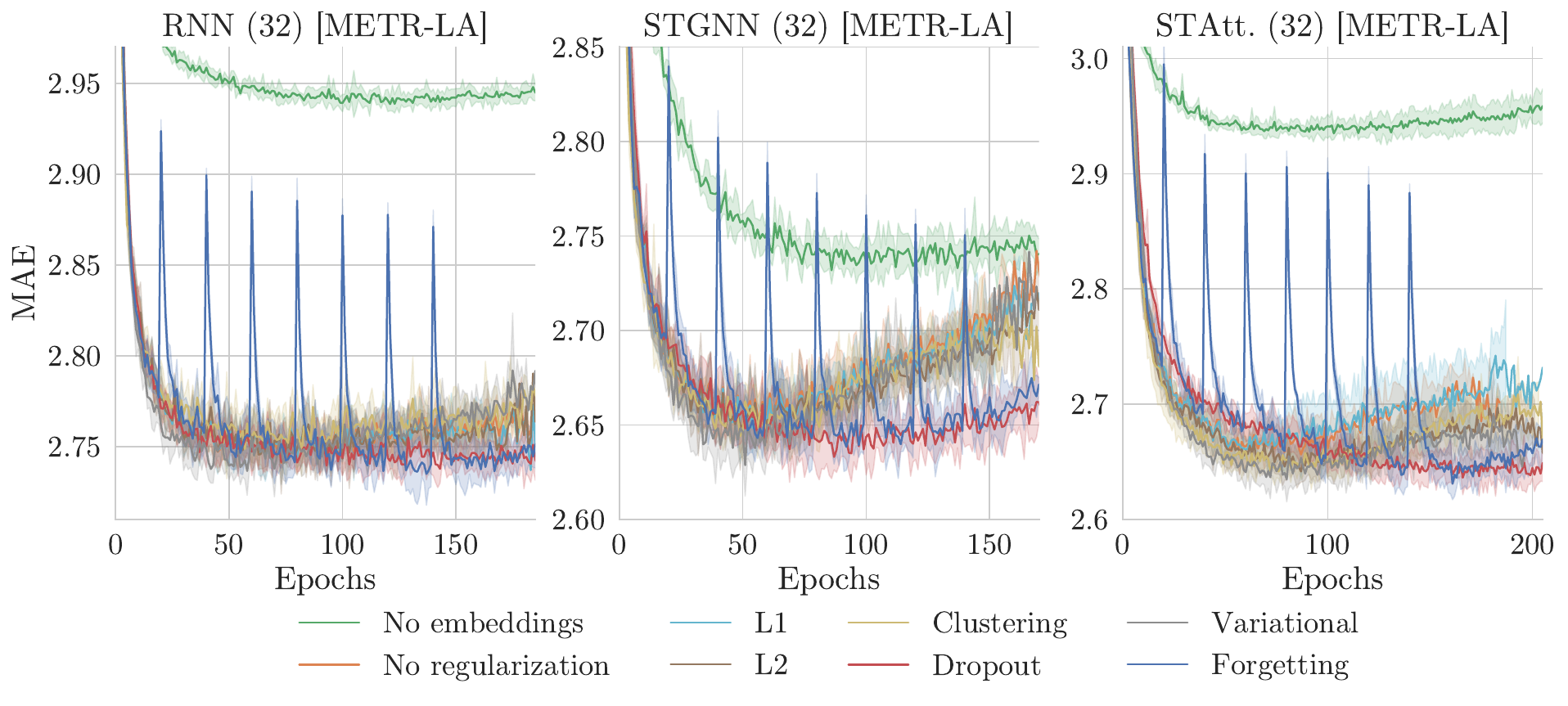}
    \vspace{-.5cm}
    \end{subfigure}
    \begin{subfigure}{0.95\textwidth}
     \centering
    \includegraphics[width=\textwidth]{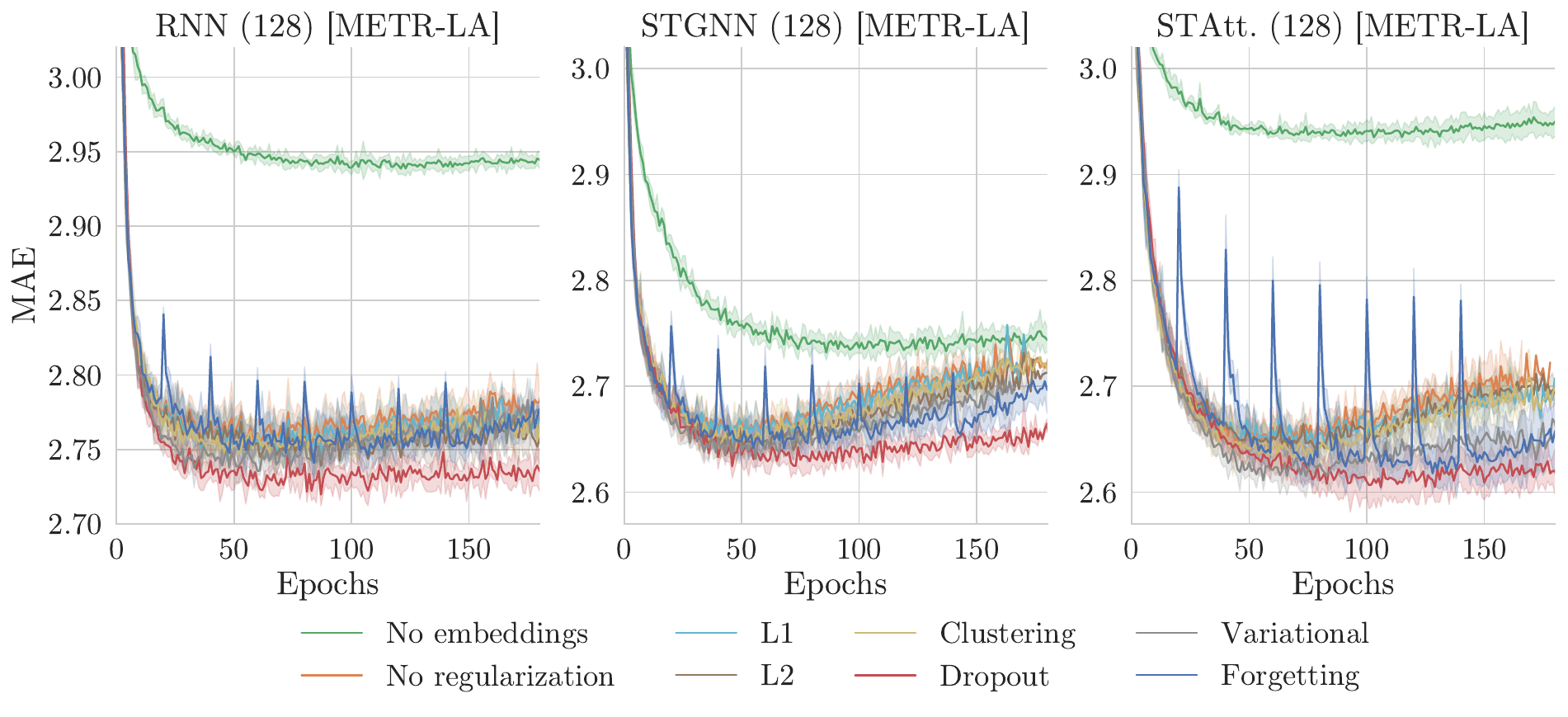}
    \vspace{-.5cm}
    \end{subfigure}
    \caption{Validation curves for different model families (5 runs, $\pm 1std$). Plot names follow the convention \textit{model (embedding size) [dataset]}.}
    \label{fig:complement_plots}
\end{figure}
Similarly, Fig.~\ref{fig:complement_plots} complements the results shown in Fig.~\ref{fig:overfit}, with the models that were not shown. We can notice a similar pattern in which dropout~(\textit{red}) can be problematic at smaller embedding sizes~(top row). Nonetheless, this seems to be less significant for the \gls{stgnn} in this specific scenario. In general, the additional plots confirm that dropout~(\textit{red}) and forgetting~(\textit{blue}) lead to different learning curves compared to the un-regularized model~(\textit{orange}), while other regularizations have little impact on this aspect.

\begin{figure}[t]
     \centering
    \includegraphics[width=\textwidth]{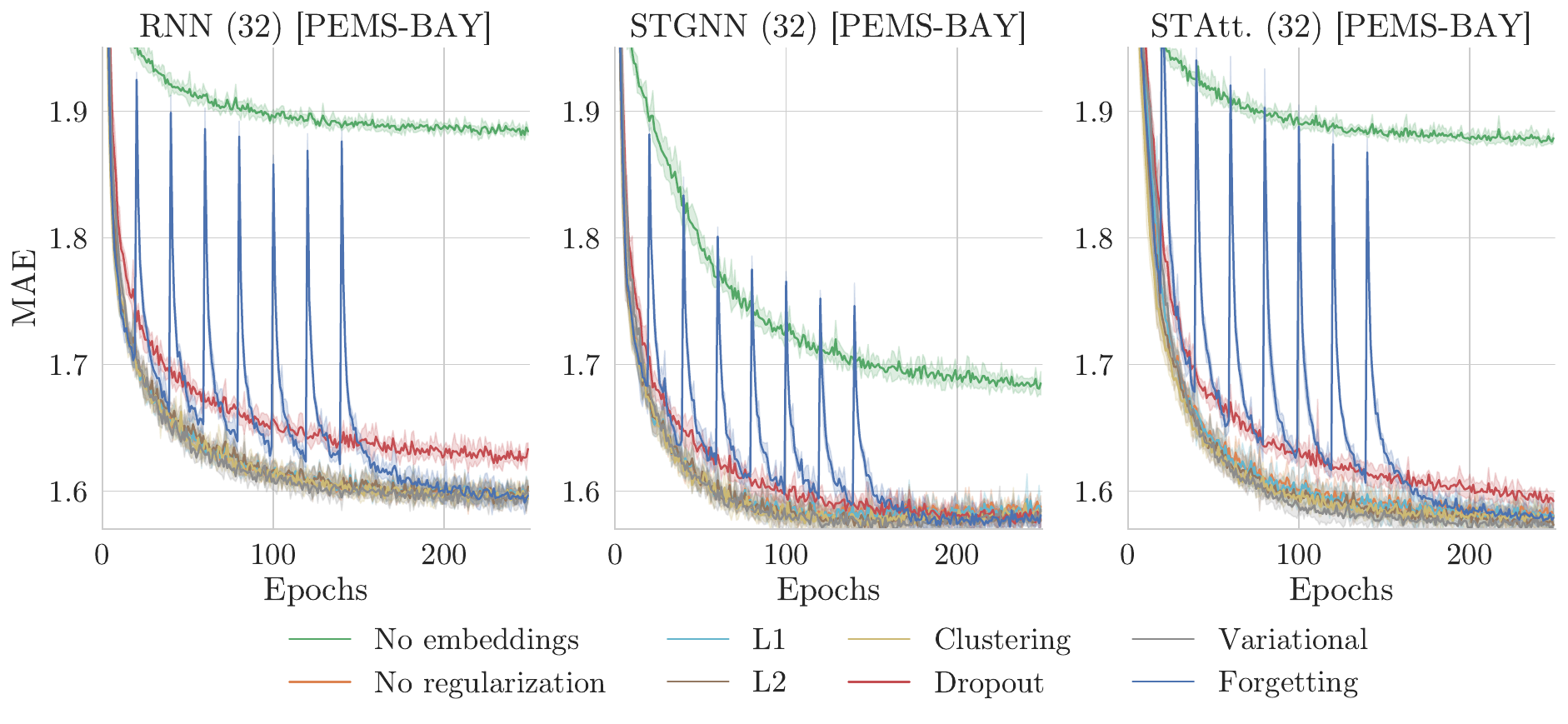}
    \vspace{-.7cm}
        \caption{Validation curves for different model families (5 runs, $\pm 1std$). Plot names follow the convention \textit{model (embedding size) [dataset]}.}
        \label{fig:complement_pems}
\end{figure}
Fig.~\ref{fig:complement_pems} shows results obtained by training models with the same setting as Sec.~\ref{subsec:overfit} on the \gls{bay} dataset. We can see that, in this case, the validation curves barely plateau. In this scenario, dropout shows potentially problematic effects, similar to those observed for small embedding sizes. 

\begin{figure}[t]
     \centering
    \includegraphics[width=\textwidth]{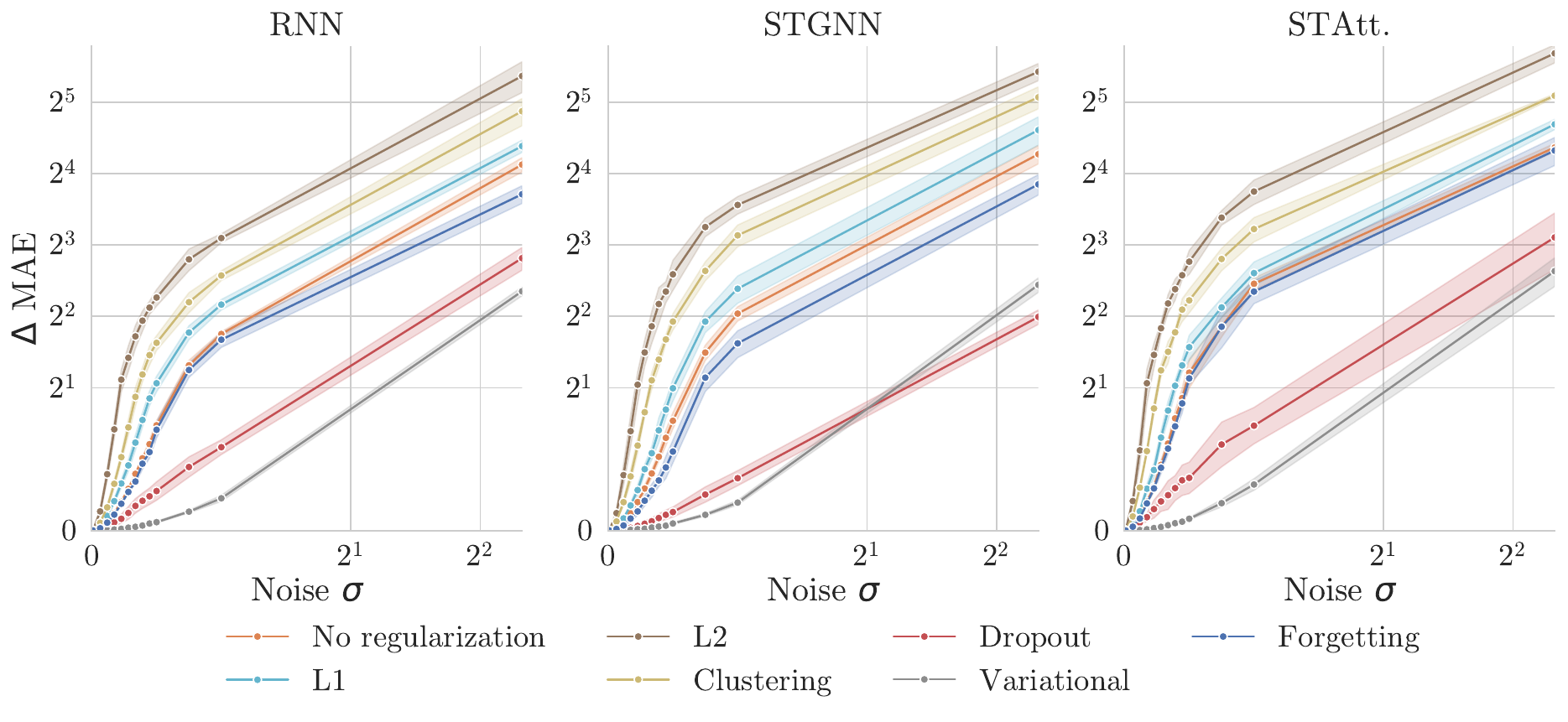}
    \vspace{-.7cm}
        \caption{Test performance degradation when adding zero-mean Gaussian noise to the embeddings, with increasing variance (5 runs, $\pm 1std$, \gls{la}). Fixed hidden size and learning rate, $64$ and $0.00075$ respectively.}
        \label{fig:app_noise_sensitivity}
\end{figure}
\begin{figure}[t]
     \centering
    \includegraphics[width=\textwidth]{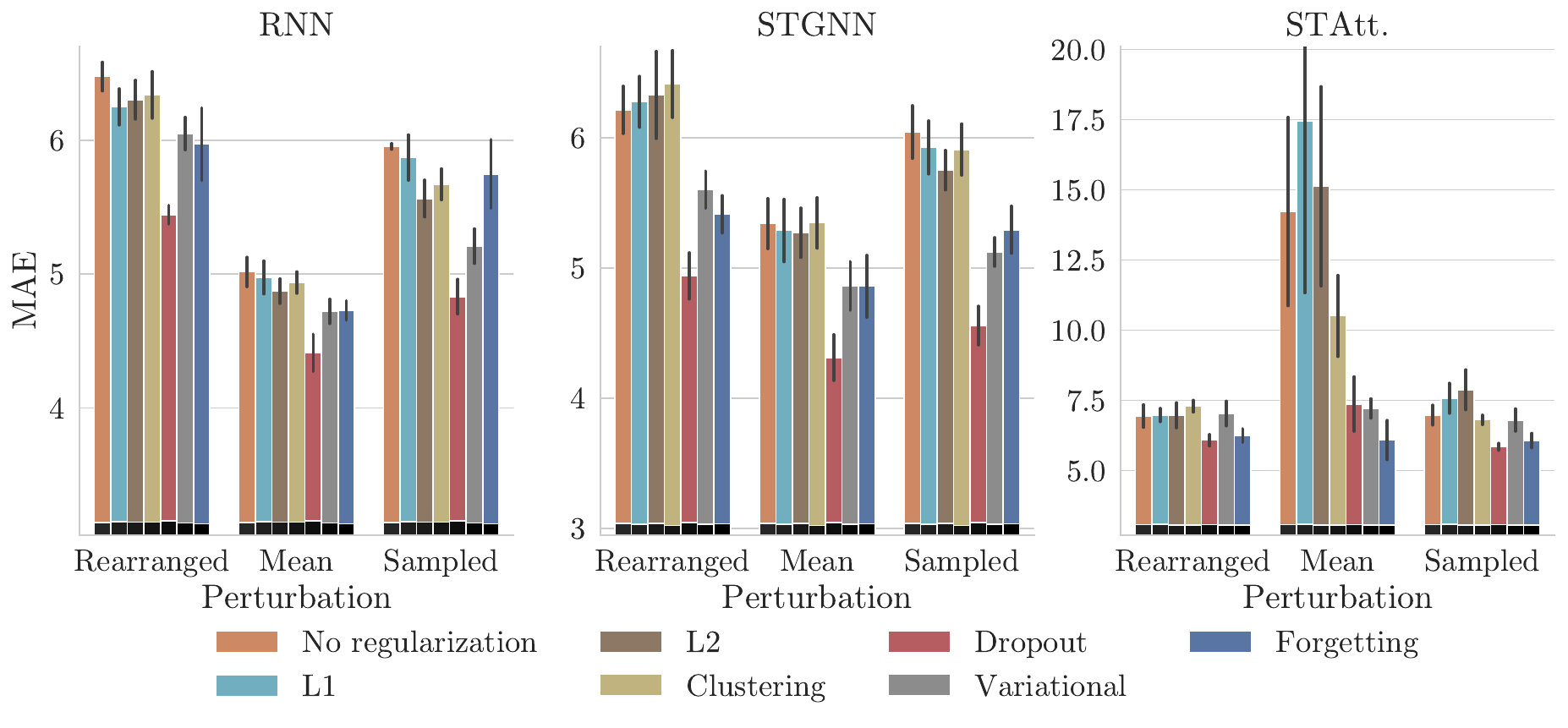}
    \vspace{-.7cm}
        \caption{Test performance degradation on embeddings perturbation (5 runs, $\pm 1std$, \gls{la}).(\textbf{Left}) random shuffling, (\textbf{Middle}) replaced with their mean, and (\textbf{Right}) replaced by a draw from their sample normal. Fixed hidden size and learning rate, $64$ and $0.00075$ respectively.}
        \label{fig:app_perturbation_sensitivity}
\end{figure}
Finally, Fig.~\ref{fig:app_noise_sensitivity} and Fig.~\ref{fig:app_perturbation_sensitivity}, provide complementary results for Fig.~\ref{fig:perturbation}. We can notice how the different regularizations rank similarly, in terms of global model robustness, across different architectures.

\end{document}